\documentclass[10pt,twocolumn,letterpaper]{article}
\usepackage{microtype}
\usepackage[accsupp]{axessibility}

\usepackage{cvpr}              % To produce the CAMERA-READY version
\newif\ifarxiv
\arxivfalse

\usepackage{xcolor}
\definecolor{cvprblue}{rgb}{0.21,0.49,0.74}
\usepackage[pagebackref,breaklinks,colorlinks,allcolors=cvprblue]{hyperref}

\def\paperID{32056} % *** Enter the Paper ID here 2453
\def\confName{CVPR}
\def\confYear{2026}

\usepackage{xcolor}
\usepackage{wrapfig}
\usepackage{soul}
\usepackage{amsmath}
\usepackage{amssymb}
\usepackage{booktabs}
\usepackage{dsfont}
\usepackage{graphicx}
\usepackage{makecell}
\usepackage{multirow}
\usepackage{pifont}% http://ctan.org/pkg/pifont
\usepackage[group-separator={,}]{siunitx}
\usepackage{tabularx}
\usepackage{caption}
\usepackage{changepage,threeparttable}
\usepackage{colortbl}
\usepackage{algpseudocode}
\usepackage{algorithm}
\usepackage{xspace}
\usepackage{diagbox} % dans le préambule

\usepackage{mathtools}
\usepackage{microtype}
\usepackage{xfrac}
\usepackage{url}

\usepackage{catchfile}
\usepackage{longtable}
\usepackage{tabu}
\usepackage{supertabular}
\usepackage{multicol}
\usepackage{enumitem}

\definecolor{yellow}{rgb}{1,1, 0.6}
\definecolor{lightyellow}{rgb}{1,1, 0.8}
\definecolor{orange}{rgb}{1, 0.8, 0.6}
\definecolor{coral}{RGB}{246,131,65}
\definecolor{pinkred}{rgb}{1, 0.6, 0.6}
\definecolor{hotpink}{RGB}{238,64,195}
\definecolor{lavender}{RGB}{207,226,243}
\definecolor{gainsboro}{RGB}{208,224,227}
\definecolor{gainsboro2}{RGB}{217,234,211}
\definecolor{blanchedalmond}{RGB}{252,229,205}

\newcommand{\camready}[1]{#1}
\newcommand{\camchange}[2]{#1}

\newcommand{\spike}{\textsc{Spike}\xspace}

\title{Learning to Solve PDEs on Neural Shape Representations}
\author{
Lilian Welschinger$^{1}$ \quad
Yilin Liu$^{1}$ \quad
Zican Wang$^{1}$ \quad
Niloy J. Mitra$^{1,2}$\\[0.3em]
$^{1}$University College London\quad
$^{2}$Adobe Research
}

\begin{document}

%\maketitle
\twocolumn[{%
\renewcommand\twocolumn[1][]{#1}%
\maketitle
\thispagestyle{empty}
\vspace*{-.2in}
\begin{center}
\centering 
\captionsetup{type=figure}
\includegraphics[width=1.0\linewidth]{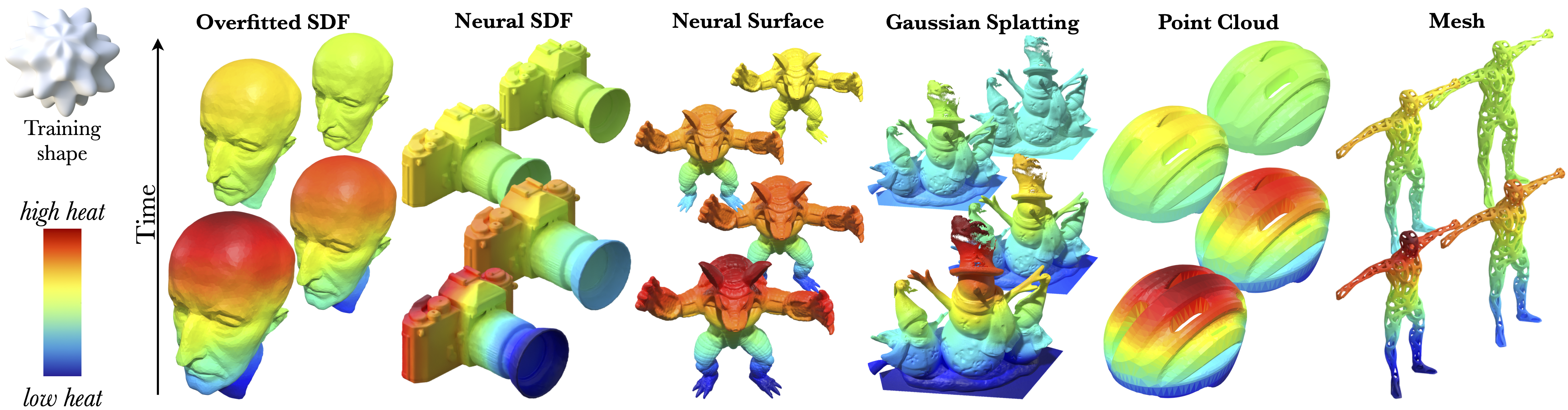}
\captionof{figure}{Our mesh-free, geometry-conditioned learned operator solves surface PDEs (\emph{heat equation} in the teaser) \emph{directly in the neural domain} on multiple modalities, without mesh extraction, or per-instance optimization. Trained once on a single exemplar (\emph{\spike}, top-left), the learned operator generalizes across unseen geometries, topologies, and input modalities. See supplemental for \camchange{detailed information.}{original: heat flow videos and also Poisson solves.} 
}%
\label{fig:teaser}%
\end{center}%
\vspace{1em}%
}]

% !TEX root = ../main.tex

\begin{abstract}
Solving partial differential equations (PDEs) on shapes underpins many shape analysis and engineering tasks; yet, prevailing PDE solvers operate on polygonal/triangle meshes while modern 3D assets increasingly live as neural representations. This mismatch leaves no suitable method to solve surface PDEs directly within the neural domain, forcing explicit mesh extraction or per-instance residual training, preventing end-to-end workflows. We present a novel, mesh-free formulation that learns a local update operator conditioned on neural (local) shape attributes, enabling surface PDEs to be solved directly where the (neural) data lives. The operator integrates naturally with prevalent neural surface representations, is trained once on a single representative shape, and generalizes across shape and topology variations, enabling accurate, fast inference without explicit meshing or per-instance optimization while preserving differentiability. 
\camready{Across analytic benchmarks (heat diffusion and Poisson equations on the sphere) and on diverse shapes and neural surface representations, our method achieves accuracy comparable to classical solvers while enabling a unified, end-to-end pipeline across neural and traditional surface representations. Our source code and project page: \small\url{https://welschinger.github.io/Learning-to-Solve-PDEs-on-Neural-Shape-Representations/}.}
% Across analytic benchmarks (heat and Poisson equations on the sphere) and real neural assets across different representations, our method slightly outperforms CPM while remaining reasonably close to FEM, and, to our knowledge, delivers the first end-to-end pipeline that solves surface PDEs on both neural and classical surface representations. 

\end{abstract}
% %%%%%%%%% BODY TEXT
% !TEX root = ../main.tex

\section{Introduction}
Solving partial differential equations (PDEs) on surfaces is central to geometry processing, shape analysis, and many engineering tasks; examples include heat flow on surfaces, Poisson equation, and harmonic interpolation. Classical FEM-based solvers~\cite{thomeeGalerkinFiniteElement2006,dziukFiniteElementMethods2013} operate on (discretized) polygonal or triangle meshes, with well-understood accuracy and stability behavior. However, they cannot directly handle contemporary 3D assets that are increasingly represented as \emph{neural shape representations} (e.g., point clouds or splats~\cite{qi2017pointnetdeeplearningpoint,zhang2023shape2vecset}, neural surfaces~\cite{morreale2021neural,williamson2025spherical}, overfitted implicit shapes~\cite{sitzmann2020siren,relufield}, neural implicit fields~\cite{park2019deepsdf,zhang2023shape2vecset}). These representations are popular as they are differentiable, often topology-agnostic, and integrate naturally with modern learning and generative systems. This creates a mismatch: \emph{mesh-centric PDE solvers do not operate in the domain where neural data lives}.

Handling surface PDEs \emph{directly in the neural shape domain} removes mesh extraction, preserves end-to-end differentiability (crucial for inverse problems and PDE priors), and handles topology changes without intermediate re-meshing or reparameterization, while avoiding round-trip errors and engineering overhead. However, current workarounds either \emph{extract a mesh} (e.g., marching cubes~\cite{diffMarchingCubes}, dual contouring~\cite{DualContouring} and their neural variants ~\cite{NeuralMarchingCubes, NeuralDualContouring}) and shuttle results back, hindering differentiable pipelines; or \emph{rely on per-instance residual training} (e.g., surface PINNs~\cite{fang2021physicsinformedneuralnetworkframework}), which generalizes poorly across shape variation.

We introduce a geometry-aware neural PDE solver that targets the surface-structure component of PDEs and operates directly inside neural representations while retaining classical-solver fidelity and applying equally to classical surfaces. Inspired by the \emph{Closest Point Method}~\cite{ruuth2008cpm}, our approach learns how surface geometry governs the extension of a surface field into a \emph{narrow band}—the core operation of embedding-based solvers—allowing PDEs to be extended to the ambient Euclidean space and solved directly on their surface representations. A \camchange{compact}{lightweight} neural operator captures local geometric context (e.g., normals and principal curvature directions) and produces this extension. Trained once on a single shape, it generalizes across unseen geometries, modalities, and topologies, requires no meshing or per-instance optimization, and remains fully differentiable \camchange{-- offering a new neural PDE layer in existing training pipelines.}{making it suitable as a new neural PDE layer into existing neural training setups.}

We demonstrate our neural PDE operator on different popular representations: meshes, spherical neural surfaces~\cite{williamson2025spherical}, point clouds~\cite{kerbl2023gaussiansplatting}, overfitted occupancy fields~\cite{mescheder2019occupancy}, Gaussian splatting~\cite{kerbl2023gaussiansplatting}, as well as deep implicit fields~\cite{park2019deepsdf}. We validate our results in two ways: 
(i)~On spheres, we evaluate the heat and Poisson equations against analytical ground truth, and compare with both FEM and CPM~\cite{ruuth2008cpm,dziukFiniteElementMethods2013}, to assess accuracy. 
% (ii)~For general shapes and modalities, we use dense-mesh FEM as reference to assess generalization and robustness.
\camready{(ii)~On general shapes and modalities, we use dense-mesh FEM as a reference, and compare with GINO \cite{li2023geometry}, while evaluating generalization across different geometries and input functions.}
Our solver achieves competitive accuracy with zero meshing overhead and no extend–restrict shuttling, as required by CPM. Most importantly, we find that our method, once trained, generalizes well across surface variations, topology, and meshing changes, owing to the local nature of the learned operator. As shown in \Cref{fig:teaser}, a model trained on a single shape (the \spike) generalizes across diverse shapes and neural representations. We also ablate design choices and hyperparameters. 

In summary, our main contributions are:
\begin{itemize}[leftmargin=*]
    \item Introducing a novel mesh-free, end-to-end differentiable solver for surface PDEs operating directly on both neural and classical surface representations.
    \item A lightweight, shape-conditioned network trained on a single shape that implicitly learns the local narrow-band extension without per-shape optimization, \camready{enabling gradient-based optimization as a differentiable PDE layer.}
    % \item a lightweight, shape-conditioned network trained on a single shape, which performs amortized grid-to-grid direct updates in a narrow band to enforce normal consistency and avoid per-shape optimization; and 
    \item Extensive evaluation showing generalization across unseen shapes, topologies, \camchange{input functions, and different shape representations}{representations, and input functions}, with competitive accuracy and speed \camchange{solving the heat diffusion}{on the heat} and Poisson equation.
\end{itemize}

\section{Related Work}

% \vspace{-0.5em}
\paragraph{Classical and mesh-based methods.}
In Euclidean domains, PDEs are classically discretized by finite differences and Galerkin finite elements~\cite{mazumderNumericalMethodsPartial2016,thomeeGalerkinFiniteElement2006}. Extending finite differences to curved manifolds typically requires embedding strategies, whereas Galerkin methods naturally generalize to arbitrary geometries via mesh-based formulations. On surfaces, \emph{surface finite elements} (SFEM) discretize the manifold and apply intrinsic schemes on a triangulation, offering strong accuracy and stability guarantees under standard regularity and shape-regular mesh assumptions; see the survey~\cite{dziukFiniteElementMethods2013}. Discrete differential geometry operators (e.g., the cotangent Laplacian) are also widely used for geometry processing and harmonic problems on meshes~\cite{meyer2003discrete,desbrun1999implicitfairing}. The main limitations are geometric and practical: performance hinges on mesh quality, evolving or noisy geometries often require (re)meshing, and distortion/tangling can degrade conditioning, accuracy, and robustness.
The main restriction being that such methods cannot directly be applied to current neural representations, without meshing.

\vspace{-1.5em}
\paragraph{Embedding and unfitted methods.}
Embedding methods solve surface PDEs in the ambient domain while enforcing surface constraints. We build on the \emph{Closest Point Method} (CPM), which alternates extension and Cartesian updates on a narrow band and is valued for simplicity and robustness~\cite{ruuth2008cpm}; accuracy/flexibility have been boosted with high-order and meshfree RBF–FD stencils and least-squares implicit variants, including moving surfaces~\cite{petras2018rbffdcpm,petras2019lsicpm}. Stochastic \emph{Projected Walk on Spheres} offers discretization-free Monte Carlo solutions via repeated manifold projections~\cite{sugimoto2024pwos}, and CPM has been adapted to interior boundaries~\cite{king2024cpmibc}. Unfitted FEM avoids explicit surface meshes by solving on a background grid: \emph{CutFEM} stabilizes cut cells with ghost penalties~\cite{burman2015cutfem,burman2025cutfemreview}, while \emph{TraceFEM} restricts spaces to an implicit level set and extends to evolving interfaces~\cite{olshanskii2016tracefem,lehrenfeld2018tracefemEvolving}. %Despite reduced meshing effort, these approaches still shuttle information between surface and grid, introducing overhead and potential bias, especially with implicit surfaces. In contrast, our method performs \emph{grid-to-grid} updates without extend–restrict loops.
\camready{While these approaches are effective and well-established, they rely on explicit surface representations and repeated interaction between surface and embedding grids. In contrast, our method learns a local, geometry-conditioned update in the narrow band, avoiding explicit extension steps.}

\vspace{-1.5em}
\paragraph{Learning-based solvers.} 

Physics-Informed Neural Networks (PINNs)~\cite{RAISSI2019686} impose PDE residuals and boundary terms in the training loss, enabling mesh-free forward/inverse solves but typically requiring \emph{per-instance} optimization; they are sensitive to stiffness, boundary enforcement, residual weighting, and training stability at scale. Surface extensions (e.g.,~\cite{fang2021physicsinformedneuralnetworkframework}) demonstrate feasibility on manifolds without meshing yet inherit the same optimization and runtime burdens. A complementary direction, \emph{neural operator} (DeepONet, FNO)~\cite{lu2021deeponet,kovachki2021neural,li2021fourier} and manifold variants, such as GINO~\cite{li2023geometry,pfaff2021learning,chen2023norm}, amortizes solution maps across problem families but generally relies on supervision from classical solvers, assumes fixed discretizations/charts, and does not natively target neural implicit geometry. Closer to our goals, \emph{implicit neural spatial representations} treat an implicit neural representation~(INR) as the spatial discretization and evolve its weights over time to solve time-dependent PDEs~\cite{chenwu2023insr}. These methods show strong accuracy–memory trade-offs but still operate via global weight evolution and per-problem time integration, and cannot be directly used to unseen shapes. 
In contrast, we learn a \emph{local, geometry-conditioned update operator} that works directly in a \emph{narrow band} around the surface, enabling generalization across shapes and representations. It takes geometric cues from diverse neural shape representations and performs a single forward update, avoiding per-instance training and mesh dependencies while retaining solver-level accuracy.

\vspace{-1.2em}
\paragraph{Neural shape representations.}
Modern 3D pipelines increasingly favor neural implicit/explicit representations over traditional meshes. Point clouds (e.g., PointNet/PointNet++ or splats~\cite{kerbl2023gaussiansplatting}) provide a mesh-free sampling interface but lack continuity and differential structure by default~\cite{qi2017pointnetdeeplearningpoint,qi2017pointnetplusplus}. Neural \emph{implicit} fields capture geometry as continuous functions: signed distance fields~(DeepSDF~\cite{park2019deepsdf}) and occupancy networks~\cite{mescheder2019occupancy} model surfaces at effectively infinite resolution and are widely used for reconstruction and analysis. Overfitted implicit neural representations, such as SIREN~\cite{sitzmann2020siren}, fit a single shape/scene as a coordinate MLP and expose smooth values and derivatives. For genus-0 surfaces, spherical neural surfaces map $\mathbb{S}^2$ to embedded shapes and expose intrinsic operators without meshing~\cite{williamson2025spherical}. Scene appearance and volume are commonly modeled by neural radiance fields (NeRF)~\cite{mildenhall2020nerf}, with real-time explicit variants via 3D Gaussian splatting~\cite{kerbl2023gaussiansplatting}. Triplane feature layouts (e.g., EG3D~\cite{chan2022eg3d}) factor 3D into three orthogonal 2D feature planes that are both expressive and efficient for reconstruction and generation. Finally, recent latent encodings for neural fields, such as 3DShape2VecSet~\cite{zhang2023shape2vecset}, represent shapes as sets of vectors tailored for generative modeling and downstream learning. These representations are differentiable and often topology-agnostic, making them suitable for our PDE solver that operates without mesh extraction (see \Cref{sec:evaluation}).

% !TEX root = ../main.tex

\section{Method}

We propose a representation-agnostic solver that computes surface PDEs 
%\emph{directly in the neural domain}, 
on \emph{neural surfaces}, while remaining compatible with other geometric representations. % Given a neural surface or field (e.g., SNS, SDF/occupancy INR, overfitted implicit, point cloud with normals), we build a narrow Cartesian band around the embedded surface and extract local geometric context~(e.g., normals, curvature tensor) at surface samples. Instead of the classical extend–solve–restrict loop in CPM~\cite{ruuth2008cpm}, we use a \emph{geometry-conditioned local operator}: a lightweight MLP that, conditioned on a small grid stencil and the local geometric context, produces a direct grid-to-grid update that advances the PDE while keeping the solution constant along implicitly defined surface normals -- the central assumption underlying embedding-based surface solvers. This decouples geometry from solving: geometry features are precomputed and reused across iterations, while the learned operator applies uniformly across shapes and topologies.
% Given a neural surface or field representation
Given a surface, which may be neural (e.g., Spherical Neural Surface, SDF or occupancy INR, overfitted implicit, or point cloud with normals), we first extract local geometric context such as normals and curvature tensor at sampled surface points. We then build a narrow Cartesian band around the embedded surface, following the principle of the \emph{Closest Point Method}~\cite{ruuth2008cpm}, and reformulate the surface PDE as a volumetric one defined within this neighborhood. Surface functions are extended to the band through a closest point extension that enforces normal constancy — the main assumption underlying embedding-based solvers. \textit{This extension is modeled by a lightweight geometry-conditioned neural operator that learns it implicitly.} The operator acts locally across the surface, recognizing the underlying geometry from local features and grid stencils to produce local band functions, which are then assembled into a single global solution. This design helps generalize across a wide range of shape modalities, topologies, and surface functions.

Our training is local and data-efficient: patches from a single representative shape (\spike in \Cref{fig:teaser}) suffice to learn the operator. Two aspects are \camchange{key}{central} to its \mbox{construction}:
% (i)~how to structure the operator architecture; (ii)~how to train the operator so that it generalizes to different functions at test time. 
% (i)~structuring the operator architecture, and
(i)~structuring \camchange{a geometric conditioned architecture}{the architecture with geometric conditioning}, and
% (ii)~how to train the operator so that it generalizes to different functions at test time. 
(ii)~ensuring \camchange{test time generalization to new functions.}{generalization to unseen functions at test time.}
% Since we express the local patches in an intrinsic coordinate framework, our operator has a simpler update rule to learn. 
% Additionally, being local patch based, the method generalizes to both unseen surfaces as well unseen functions on the surface. 
%Additionally, being local, the method generalizes to both unseen surfaces as well unseen functions on the surface.
% At inference, the method reduces to repeated forward passes of the local operator for time-dependent PDEs, requires no meshing or per-instance optimization, remains fully differentiable, and integrates as a drop-in neural PDE layer. We start by recapping the CPM method. 
At inference, \camchange{it}{the method first} produces the extended band function by \mbox{applying} the local operator across all patches — each acting locally but contributing to a single global update of the field. The PDE is then solved directly within the band, as in~\cite{ruuth2008cpm}, and for time-dependent problems, this process reduces to repeated global updates over time steps. The approach requires no meshing or per-instance optimization, remains fully differentiable, and integrates as a drop-in neural PDE layer.
%We start by recapping the Closest Point Method.

\begin{figure*}[t!]
    \centering
\includegraphics[width=1\linewidth]{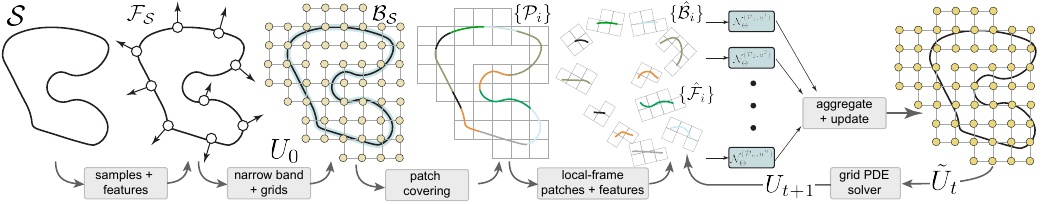}
    \caption{
    \textbf{Pipeline overview.} 
From a surface $\mathcal{S}$, we sample points and geometric features (normals, local features). 
Around an $\varepsilon$-\emph{narrow band} around the shape, we gather Cartesian grids $\mathcal{B}_{\mathcal{S}}$ to store an initial field $U_0$ extended from surface values and covered by overlapping, surface-centred patches $\{\mathcal{P}_i\}$. 
Each patch is reoriented to its local frame, yielding $\{\hat{\mathcal{B}}_i,\hat{\mathcal{F}}_i\}$, which are processed by our lightweight geometry-conditioned operators $\mathcal{N}_\Theta$ to produce local updates (see \Cref{fig:SUN_network}).  
The local updates are smoothly aggregated to form the global band update $\tilde{U}_{t}$ and advanced with a standard grid PDE time step to get $U_{t+1}$. 
Iterating this loop yields surface PDE solutions directly in the neural domain via grid-to-grid updates, \textit{without} mesh extraction or extend–restrict shuttling.
}
\vspace*{-.1in}
    \label{fig:overall_method}
\end{figure*}

\vspace{-1em}
\paragraph{Closest Point Method (CPM).}
The original method~\cite{ruuth2008cpm}  embeds surface PDEs in a thin Cartesian \emph{narrow band} around the surface $\mathcal{S}$. The equivalent volumetric PDE is then solved inside this band using standard numerical methods such as finite differences (FD), Runge–Kutta~(RK) schemes and time integrators such as forward Euler for time-dependent problems. For the solution of the volumetric PDE to coincide with that of the original surface PDE when restricted to~$\mathcal{S}$, \camchange{its gradient along normals are zero}{}, the surface function must be extended into the band through a \emph{closest point extension}, ensuring constancy along surface normals.

% The main idea is to perform a \textit{closest point extension} of the functions defined on the surface to a neighborhood around it. 
% For a suitable \textit{extension}, solving the surface PDE amounts to solving the easier PDE in this volumetric neighborhood using standard numerical methods such as finite differences (Euler solve) or Runge–Kutta~(RK) schemes.

The CPM alternates between two simple operations: (i)~a standard discrete volumetric \emph{solve} in the \emph{narrow band}, and (ii)~a \emph{re–extension} step to ensure normal constancy. At the core is the closest–point map $\mathrm{cp}$ sending a band point $x\in\mathbb{R}^3$ to its nearest point on the surface $\mathrm{cp}(x)$. A surface function $u_{\mathcal{S}}$ is extended to the band by  $u(x):=u_{\mathcal{S}}(\mathrm{cp}(x))$. When $u$ is constant along normals, ambient derivatives restricted to surface $\mathcal{S}$ coincide with intrinsic ones, allowing replacement of surface operators (e.g., gradient/Laplacian) by standard finite–difference stencils during \emph{solve} step. 

In classical CPM, surface PDEs are solved by combining Cartesian updates in a narrow band with extension steps that maintain consistency of the solution along surface normals. In practice, CPM requires (i) constructing a band of width $\varepsilon$, (ii) efficient evaluation of closest points, and (iii) \emph{re-extension} operation, and (iv) consistent handling of boundary conditions by tagging band cells whose projections lie on any boundary $\partial\mathcal{S}$. For details,  see the original paper \cite{ruuth2008cpm}. 

%The CPM is simple, robust, and reuses off–the–shelf Cartesian solvers; however, it requires \emph{extend–solve–reextend}, using surface information. We instead learn this step as a local, geometry-conditioned operator, enabling generalization across shapes and representations, including neural modalities.

\camready{The CPM provides a simple and robust framework for solving surface PDEs using Cartesian discretizations. We extend \camchange{it}{this paradigm} to operate across diverse surface representations, including neural ones, by learning a geometry-conditioned local operator that performs the extension implicitly.}

%%%%%%%%%%%%%%%%%%%%%%%%%%%%%%

\subsection{Algorithm Steps}
\label{sec:algorithm}
We learn a local neural solver that operates around a \emph{narrow band} of implicit or explicit surface and produces band updates that keep the function constant along surface normals. Our pipeline is representation-agnostic and applies to any surface that supports (i) sampling on the surface,
%or projecting onto the surface, 
and (ii) estimating local differential cues. 
In \Cref{sec:evaluation}, we discuss how to estimate these for different neural shape representations.

\vspace{-1em}
\paragraph{Inputs and notation.}
Let $\mathcal{S} \subset \mathbb{R}^3$ be a smooth surface representation, either neural (e.g., SNS, SDF/occupancy INR) or traditional (e.g., point cloud, mesh).
% Let $\mathcal{S}\subset\mathbb{R}^3$ be a smooth surface given by a neural representation (e.g., SNS, SDF/occupancy INR). 
We denote the unit normal at $x\in\mathcal{S}$ by $\mathbf{n}(x)$, and the principal curvature directions by $(\mathbf{t}_1(x),\mathbf{t}_2(x))$, with $\mathbf{t}_1$ aligned to the maximum curvature and $\mathbf{t}_2$ to the minimum; when curvatures are equal (umbilic points), any orthogonal tangent pair is chosen \camchange{regardless of the order}{and the order doesn't matter}. A regular Cartesian grid $G$ provides samples in a narrow $\varepsilon$-band around $\mathcal{S}$. 
\camchange{Full pipeline in \Cref{fig:overall_method}.}{
We now describe our full pipeline (see \Cref{fig:overall_method}).}

\vspace{-1em}
\paragraph{Pipeline Overview.}
(i) \emph{Surface feature extraction.}
Compute geometric descriptors on the surface on a set of surface samples as:
$
\mathcal{F}_{\mathcal{S}}:=\big\{(x,\; \mathbf{n}(x),\; \mathbf{t}_1(x),\; \mathbf{t}_2(x))\ \big|\ x\in\mathcal{S}\big\}$.

(ii) \emph{Narrow-band construction.}
Define a uniform volumetric grid $G$ in a bounding box of $\mathcal{S}$ and retain grid nodes within distance $\varepsilon$ of the surface:\mbox{
$
\mathcal{B}_{\mathcal{S}}:=\big\{y\in G\ \big|\ \mathrm{dist}(y,\mathcal{S})\le \varepsilon\big\}.
$}
%We also cache the projection $\Pi(y)$ and geometric features at each band site $y\in\mathcal{B}_{\mathcal{S},\varepsilon}$. %\lilianc{I didn't understand the last sentences. We don't need the closest point in our method. The only time we needed it was for a training to have a ground truth}
%\lilianc{Should we include epsilon in the notation? This lead to heavy notations no?}\robert{should we mention grid size here? like C * g instead of epsilon?}

(iii) \emph{Overlapping local patches.}
Cover $\mathcal{S}$ and its \emph{narrow band} $\mathcal{B}_{\mathcal{S}}$ with surface-centered patches. Each patch is anchored at a surface point $p_i^c \in \mathcal{S}$, which serves as its center, and is defined as:
$
\mathcal{P}_i:= (\mathcal{L}_i, \ \mathcal{B}_i,\ \mathcal{F}_i),
$
where $\mathcal{L}_i = (p_i^c, \mathbf{n}(p_i^c), \mathbf{t}_1(p_i^c), \mathbf{t}_2(p_i^c))$ defines a local frame at $p_i^c$ used to express all subsequent quantities. $\mathcal{B}_i\subset\mathcal{B}_{\mathcal{S}}$ collects nearby band samples and $\mathcal{F}_i$ gathers surrounding surface features (points and normals in our implementation).
% where $\mathcal{L}_i = (p_i^c, \mathbf{n}(p_i^c), \mathbf{t}_1(p_i^c), \mathbf{t}_2(p_i^c))$ denotes the surface feature at the patch center $p_i$; while $\mathcal{B}_i\subset\mathcal{B}_{\mathcal{S}}$ collects nearby band samples, and $\mathcal{F}_i$ gathers surrounding surface features (points and normals). %The union of $\{\mathcal{B}_i\}$ covers $\mathcal{B}_{\mathcal{S}}$. \lilianc{the last sentence is redundant as we already said that we want to cover the band}
% Note that each patch is expressed in the local coordinate system of its center $p_i$ -- thus the patches, along with their local grids, get recentered and repositioned based on the intrinsic coordinate frame of $p_i$. 
All quantities within a patch are expressed in the local frame $\mathcal{L}_i$. Accordingly, coordinates of local band samples and local surface features (i.e., $\mathcal{B}_i$, $\mathcal{F}_i$) are transformed into this intrinsic coordinate system, and we denote their local-frame representations with a hat symbol (i.e., $\hat{\mathcal{B}}_i$, $\hat{\mathcal{F}}_i$).

(iv)~\emph{Learned band-to-band update (neural operator).}
Given a scalar band field $U_t:\mathcal{B}_{\mathcal{S}}\to\mathbb{R}$ at time $t$, we denote by $u_t^{i} := U_t|_{\mathcal{B}_i}$ its restriction to the local band associated with patch $\mathcal{P}_i$. A lightweight neural operator $\mathcal{N}_\Theta$ consumes per-patch stencils (band values and local geometry) and updates the sampled field to enforce normal constancy:
$
\tilde{u}_t^{i} = \mathcal{N}_\Theta^{(\mathcal{P}_i, u^i_t)} (\hat{\mathcal{B}}_i). 
$
We denote by a tilde $(\tilde{\cdot})$ functions that are constant along surface normals.

(v)~\emph{Aggregation of local predictions.}
Local predictions are combined to reconstruct a global band field via smooth, proximity-weighted averaging as, 
\begin{equation*}
    \tilde{U}_t(x) =
    \frac{\sum_{i: x \in \mathcal{B}_i} \exp(-\|x - p_i^c\|^2 / T)\, \tilde{u}_t^i(x)}
         {\sum_{i: x \in \mathcal{B}_i} \exp(-\|x - p_i^c\|^2 / T)}, 
\end{equation*}
with temperature $T>0$ controlling the blending. 

(vi) \emph{Time evolution in the band.}
Since $\tilde{U}_t$ is constant along normals, intrinsic surface operators can directly be accurately approximated. We evolve $\tilde{U}_t$ using standard finite differences and forward Euler scheme (e.g., for heat/diffusion):
\[
U_{t+dt}(x) = \tilde{U}_t(x)+dt\,\Delta \tilde{U}_t(x),\qquad x\in\mathcal{B}_{\mathcal{S}},
\]
where $\Delta$ is the discrete Laplacian on $G$ restricted to the band. Boundary conditions are imposed on band nodes whose projections lie on $\partial\mathcal{S}$, as in \cite{ruuth2008cpm}. Note that $U_{t+dt}$ carries no tilde, as there is no guarantee that the updated function remains constant along the surface normals.

(vii) \emph{Iterate or reconstruct.}
If additional steps are needed, return to the \emph{learned band-to-band update} (step~iv) and repeat the aggregate–evolve cycle. At any time, we `readout' the surface solution by restricting the band field to $\mathcal{S}$, by interpolating with radial basis functions (Gaussian kernels in ours) for a smooth surface field.
% e.g., $u(x)=U_T(\Pi(x))$, and (optionally) interpolate with radial basis functions (Gaussian kernels in ours) for a smooth surface field.
%\end{enumerate}

%%%%%%%%%%%%%%%%%%%%%%%%%%%%%%

\subsection{Overlapping Local Patches}
\label{patch_related}

We decompose the surface $\mathcal{S}$ and its \emph{narrow band} $\mathcal{B}_\mathcal{S}$ into overlapping, surface–centered local patches, each aggregating nearby band samples for grid-based updates and nearby surface samples with geometric features for conditioning. A patch $\mathcal{P}_i$ is centered at a surface point $p_i^c\in\mathcal{S}$. Around $p^c_i$, we gather the $k$ nearest band nodes to form a local stencil $\mathcal{B}_i$. In \Cref{sec:evaluation}, we discuss choice of $k$ for good accuracy–locality trade-off. Next, we take tight axis-aligned bounding box of $\mathcal{B}_i$ and dilate it by a small margin. All surface samples whose coordinates lie inside this enlarged box, together with their normal, constitute the surface-conditioning set $\mathcal{F}_i$, providing a broader geometric context around the local surface region. Thus, the tuple $\mathcal{P}_i:= (\mathcal{L}_i,\mathcal{B}_i,\mathcal{F}_i)$, with $\mathcal{L}_i = (p_i^c, \mathbf{n}(p_i^c), \mathbf{t}_1(p_i^c), \mathbf{t}_2(p_i^c))$, defines one such patch. Using $\mathcal{L}_i$, we express all quantities of $\mathcal{B}_i$ and $\mathcal{F}_i$ in the local frame centered at $p_i^c$ with basis $(\mathbf{n}(p_i^c), \mathbf{t}_1(p_i^c), \mathbf{t}_2(p_i^c))$, ensuring invariance to translation and rotation. Degenerate cases where curvature directions are ambiguous 
(e.g., umbilic regions where principal curvatures coincide) are naturally present in the training data and are further handled  through data augmentation: random rotations of the local patch enforce the network to learn rotational invariance.

We progressively generate patches across $\mathcal{S}$, expanding outward from an initial (surface) seed so that coverage naturally propagates over the surface (similar to floodfill restricted to the surface). This strategy yields a family of overlapping patches whose union covers the entire band, while maintaining controllable redundancy. The degree of overlap is controlled by a spacing parameter in our patch-placement procedure, which determines how far each new center is placed from the previous ones while ensuring that adjacent band regions still overlap. Smaller spacing increases redundancy and overlap \camchange{improving robustness but adds computational cost, whereas larger spacing yields sparser coverage.}{, whereas larger spacing yields sparser coverage. Increasing either improves robustness but adds computational cost.} 

\vspace{-1em}
\paragraph{Coverage condition.}
A potential issue arises from using a fixed number of neighbours $k$ to define each patch: 
for small grid spacing $\Delta x$ or large band width $\varepsilon$, 
some band points may lie too far from any surface center $p_i^c$, 
leading to incomplete coverage of the band $\mathcal{B}_\mathcal{S}$. 
This motivates us to seek a relation linking $\varepsilon$, $\Delta x$, and $k$.

This setting is closely related to the classical \emph{Gauss circle problem} 
(and its three-dimensional analogue, the \emph{Gauss sphere problem}, see \cite{Kratzel1988LatticePoints}), 
which counts the number of lattice points contained in a ball of radius $r$. 
In three dimensions, neglecting higher-order terms, the number of grid points $N_3$ within a band of radius $\varepsilon$ and spacing $\Delta x$ is well approximated by the volume of a ball of radius ${\varepsilon}/{\Delta x}$: 
\[
N_3\!\left({\varepsilon}/{\Delta x}\right) \approx 
\tfrac{4}{3}\pi \left(\frac{\varepsilon}{\Delta x}\right)^{\!3}.
\]
Ensuring every band point is covered by at least one patch yields the condition, we arrive at: 
\[
\varepsilon \le \Delta x \left(\tfrac{3k}{4\pi}\right)^{1/3}.
\]
\camready{In practice, we set $\varepsilon$ to a small fraction of the shape size (e.g., $\sim$5\%) and choose $\Delta x$ to satisfy this coverage condition, balancing accuracy and computational cost.}

\subsection{Learning a Neural Update Operator}

\subsubsection{Neural geometry encoder}
Our neural geometry encoder is a lightweight network built from small MLPs with an attention-like interaction for local geometry adherence. It operates locally in the \emph{narrow band} around the surface and updates the target function in a single step, \textit{directly} on grid values.

\begin{figure}[t!]
    \centering
    \includegraphics[width=\linewidth]{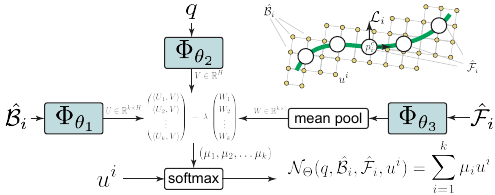}
    \caption{
    \textbf{Neural update operator (overview).} 
Given a query location $q$, the local band $\hat{\mathcal{B}}_i$ expressed in the local frame $\mathcal{L}_i$ centered around $p_i^c$ for patch $\mathcal{P}_i$, with locally-transformed surface features $\hat{\mathcal{F}}_i$ (e.g., positions, normals), and current band values $u^i$ at grid sites (time index $t$ omitted for brevity; full notation $u^i_t$), our operator predicts updated function value at location $q$. 
Trainable components include compact MLP blocks ($\Phi_{\theta_1}, \Phi_{\theta_2}, \Phi_{\theta_3}$) and a scalar $\lambda$. 
The full network $\mathcal{N}_\Theta$ produces the updated band value at $q$, yielding a single geometry–conditioned grid-to-grid step.
    } 
    \label{fig:SUN_network}
\end{figure}

\vspace{-1em}\paragraph{Network architecture.}
\label{sec:model_architecture}
As illustrated in \Cref{fig:SUN_network}, our neural solver acts on local geometry attributes and updates the field within each narrow band $\mathcal{B}_i$. The inputs are the query point, the band points $\hat{\mathcal{B}}_i$, and their current field values $u^i$, together with the associated surface features $\hat{\mathcal{F}}_i$, all expressed in the local frame $\mathcal{L}_i$. The query point $q$ attends to its neighboring band samples to obtain spatial weights, while $\hat{\mathcal{F}}_i$ modulates this aggregation so the update is \textit{conditioned} on local geometry. The output is a weighted sum producing the updated value at $q$. To handle a variable number of surface features per patch, we apply mean pooling (similar to \cite{qi2017pointnetdeeplearningpoint}), yielding a fixed-size descriptor.
Formally, \camchange{the update $\mathcal{N}_\Theta$ is}{we encode the neural update as}: 
\begin{equation*}
\begin{array}{rcl}
\mathcal{N}_\Theta :
\mathbb{R}^3 \times \mathbb{R}^{k \times 3} \times \mathbb{R}^{N_i \times 6} \times \mathbb{R}^k
& \longrightarrow & \mathbb{R}, \\[6pt]
(q, \hat{\mathcal{B}}_i, \hat{\mathcal{F}}_i, u^i)
& \longmapsto &
\mathcal{N}_\Theta(q, \hat{\mathcal{B}}_i, \hat{\mathcal{F}}_i, u^i)
\end{array}
\end{equation*}
where $\Theta$ contains the weights of the three MLPs encoder $(\theta_1, \theta_2, \theta_3)$, a learnable scalar $\lambda$ and $N_i$ denotes the number of surface samples in patch $\mathcal{P}_i$, which may vary across patches. For convenience, we define a compressed form of the network where the patch $\mathcal{P}_i = (\mathcal{L}_i, {\mathcal{B}}_i, {\mathcal{F}}_i)$ and its current field $u^i$ are fixed: 
\[\mathcal{N}_\Theta^{(\mathcal{P}_i, u^i)}:q\longmapsto\mathcal{N}_\Theta (q, \hat{\mathcal{B}_i}, \hat{\mathcal{F}_i}, u^i)\]
In practice, our implementation evaluates multiple queries simultaneously. Let $Q = \left\{q_1, \cdots, q_b\right\} \subset \mathbb{R}^3$; then $\mathcal{N}_\Theta^{(\mathcal{P}_i, u^i)}(Q)=\left\{\mathcal{N}_\Theta^{(\mathcal{P}_i, u^i)}(q)\right\}_{q\in Q}$. When solving PDE, we set $Q = \hat{\mathcal{B}}_i$ as band values are updated at band samples.

\camready{This design leverages the locality of embedding \mbox{methods}: the extension step depends only on local geometry and nearby samples, making it \camchange{suitable}{well-suited} for a learned, patch-based operator that generalizes across shapes and representations.}

\subsubsection{Training setup}

\paragraph{Dataset construction.}
We train on a single representative surface, the \spike, represented by an SNS~\cite{williamson2025spherical} $S_\omega:\mathbb{S}^2\subset\mathbb{R}^3\!\rightarrow\!\spike$. 
Note that since our network only depends on first and second order quantities (i.e., normal and curvatures), the single \spike shape, having a good distribution of curvature profiles, is sufficient to train on -- we test its generalization behavior in \cref{sec:evaluation}. 
Surface is split into patches $\{\mathcal{P}_i\}$, as  \cref{patch_related}. For each patch, we apply random rotations. Each band $\mathcal{B}_i$ is paired with closest points as, 
\[
\operatorname{cp}(x):=S_\omega\Big(\arg\min_{y\in\mathbb{S}^2}\|x-S_\omega(y)\|_2^2\Big),\quad x\in\mathcal{B}_i,
\]
forming $\Pi_i:=\{\operatorname{cp}(x)\mid x\in\mathcal{B}_i\}$. We use monomials for supervision:
\[
\mathcal{M}:=\{\,(x,y,z) \mapsto x^iy^jz^k\mid i{+}j{+}k\le 5\,\}.
\]
Note that we rely on any (unseen) function to be sufficiently approximated by only their top few Taylor coefficients since PDE solutions are smooth; hence, learning over monomial functions turns out to be sufficient in our tests. 
For each such $g\in\mathcal{M}$, we evaluate $g(\mathcal{B}_i)$ and $g(\Pi_i)$, where the first serves as the network input and the second as the ground truth target, yielding pairs:
\[\mathcal{E}
_i := \big\{\, \big(g(\mathcal{B}_i),\, g(\Pi_i)\big) \;\big|\; g \in \mathcal{M} \,\big\}.\]
Thus, the training dataset is $\mathcal{D}:=\{\,(\hat{\mathcal{B}}_i,\hat{\mathcal{F}}_i,\mathcal{E}_i)\,\}_{i=1}^p$.

\vspace{-1em}
\paragraph{Loss functions.}
We use two loss functions. The primary mean squared error enforces accurate function reconstruction:
\[L_{\mathrm{MSE}}=\frac{1}{k|\mathcal{D}|| \mathcal{M}|}\!\!\sum_{\substack{(\hat{\mathcal{B}}_i,\hat{\mathcal{F}}_i,\mathcal{E}_i)\in\mathcal{D} \\(g,g^{\mathrm{GT}})\in \mathcal{E}
_i}}\!\!\,\big\|\mathcal{N}_\Theta^{(\mathcal{P}_i, g)}(\hat{\mathcal{B}}_i)-g^{\mathrm{GT}}\big\|_2^2.\]
% \[\mathcal{L}_{\mathrm{MSE}}:=\mathbb{E}_{(\mathcal{B},\mathcal{F},(g,g^{\mathrm{GT}}))}\!\left(\|\mathcal{N}_\Theta(\mathcal{B},\mathcal{F},g)-g^{\mathrm{GT}}\|_2^2\right).\]
To check geometric consistency, we monitor a normal-consistency term enforcing constancy along surface normals, evaluated over different patches and functions $(\mathcal{P}, u)$:
\[
L_{\mathrm{NC}}:=\sum_q \big|\langle \nabla_q \: \mathcal{N}_\Theta^{(\mathcal{P}, u)}(q),\, \mathbf{n}(\operatorname{cp}(q))\rangle\big|.
\]
This term encourages the field gradient to remain orthogonal to the surface normals; when the dot products are close to zero then the field is nearly constant along the surface normals. The gradient $\nabla_q \:\mathcal{N}_\Theta^{(\mathcal{P},u)}(q)$ is computed via automatic differentiation with respect to the query location $q$ (other terms are detached).
The overall objective is:
\[
L=\,L_{\mathrm{MSE}}+\alpha\,L_{\mathrm{NC}}.
\]
We conduct an ablation study  (see supplemental) to assess the contribution of the normal-consistency term $L_{\mathrm{NC}}$ and to determine an appropriate weighting $\alpha$. This analysis highlights how enforcing normal-aligned consistency improves accuracy across surfaces.
% In practice, we first train with only $\mathcal{L}_{\mathrm{MSE}}$ and monitor $\mathcal{L}_{\mathrm{NC}}$; if it decreases in tandem, the additional term is omitted. Otherwise, we include $\mathcal{L}_{\mathrm{NC}}$ to steer the model toward normal-aligned consistency.

\section{Evaluation}
\label{sec:evaluation}

%\subsection{Setup}

\paragraph{Baselines.}
We evaluate against \camready{three} established families:
(i)~\emph{Surface FEM (SFEM)} discretizes the PDE intrinsically on an explicit triangle mesh: unless stated otherwise, we use linear elements with the cotangent Laplacian and a consistent mass matrix~\cite{dziukFiniteElementMethods2013,thomeeGalerkinFiniteElement2006}. Meshes come either from the ground-truth surface or from marching cubes~\cite{10.1145/37402.37422} on the same implicit surface used by our method, with multiple resolutions to probe convergence and mesh-quality effects. We do not aim to outperform FEM, whose solvers are highly mature and extensively optimized. Instead, we use FEM to provide reliable reference solutions and expected error levels, illustrating the behavior and capability of our method. Comparisons to FEM should therefore be viewed as grounding rather than competition.
\newline
(ii)~\emph{Closest Point Method (CPM)} solves in an Eulerian \emph{narrow band} around the surface using standard Cartesian stencils and polynomial interpolation for the \emph{re-extension} step, alternating \emph{solve} and \emph{re–extend} steps~\cite{ruuth2008cpm}. Closest-point projections and normals are computed from the same implicit geometry used by our method to ensure parity. \camready{Both methods share the same solvers (explicit Euler with $\Delta t=0.1\,\Delta x^2$ for heat, sparse solver on the grid), so differences stem from the extension operator.}
\newline
(iii)~\emph{Geometry-Informed Neural Operator (GINO)} \camready{is a geometry-aware FNO variant for PDE learning on varying domains. It encodes the input shape from a point-cloud representation onto a latent grid with a GNO layer \cite{li2020neural}, processes the grid with an FNO using concatenated SDF features, and maps the prediction back to the original geometry via GNO. To compare generalizability, robustness, and efficiency, we trained GINO on two datasets: a dataset of \textbf{ShapeNet} geometries and the high-resolution spike shape used to train our model. For both datasets, ground-truth solutions were generated using a high-resolution FEM solver. Since GINO is trained separately for each PDE setting, each new PDE requires retraining, which takes approximately 8 hours.}

For fairness, all baselines share the same right-hand sides, initial data, and boundary conditions; we also matched resolution schedules and aligned stopping criteria  (final time or steady-state residual).

\vspace{-1em}
\paragraph{PDEs.} 
We benchmark \emph{heat diffusion} ($\partial_t u=\Delta_{\mathcal{S}} u$) and \emph{Poisson} ($\Delta_{\mathcal{S}} u=f$) on closed surfaces. For well-posedness, we initialize the heat equation with a prescribed initial condition and let it evolve until reaching a steady state. For Poisson, we choose a zero-mean function $f$ over the surface and select the zero-mean solution on the surface, since the kernel of the Laplace–Beltrami operator on closed surfaces corresponds to constant functions.
We report both boundary-free cases and settings with Dirichlet conditions on embedded curves (imposed identically for all methods).
% For heat, we use matched initial data and fixed horizons; for Poisson, steady-state solutions are compared under the same $f$ and normalization.

\vspace{-1em}
\paragraph{Metrics.} We report normalized mean absolute error (NMAE), normalized max error (NMaxE), normalized root mean square error (NMRSE). Input probe function ranges were normalized to $[-0.5, 0.5]$. See supplemental for details. 
%We report normalized errors and wall-clock time. For ours, we do not include training time as training is done once, and not repeated for unseen shapes. 

\vspace{-2mm}
\paragraph{Shape representations.} We evaluate across common shape encodings and derive the geometric cues needed by our operator in a consistent way.
(i)~\emph{Meshes:} normals are area–weighted averages of incident face normals, principal curvatures are obtained from discrete differential operators~\cite{meyer2003discrete,desbrun1999implicitfairing}. 
(ii)~\emph{Point clouds:} we estimate normals via PCA of $k$-NN neighborhoods with sign disambiguation along a coarse viewpoint field. More advanced point normal prediction~\cite{GuerreroEtAl:PCPNet:EG:2018} may be used. 
(iii)~\emph{Spherical Neural Surfaces~(SNS):} The method provides direct access to normals and first/second fundamental forms by differentiating the mapping $S_\omega:\mathbb{S}^2\subset \mathbb{R}^3\!\to\!\mathbb{R}^3$; principal directions of curvature follow from the Weingarten map~\cite{williamson2025spherical}. 
(iv)~\emph{Implicit SDF fields (DeepSDF/overfitted INRs):} normals are $\nabla\phi/\|\nabla\phi\|$ for these implicit fields~\cite{park2019deepsdf}. 
% \yl{However, although we could have used curvatures using level-set formulas~\cite{osher2003levelset}, we found the estimates to be noisy; hence, we did not use curvature estimates in these cases}. \lilian{maybe only mention how to compute the principal direction of curvatures} For occupancy  fields~\cite{mescheder2019occupancy,sitzmann2020siren}, we compute normals from the implicit gradient of the network near the isosurface. 
%We also obtain the principal directions by first evaluating the gradient and Hessian of the implicit function at the query position to determine the surface normal and local 3D curvature, and then projecting that Hessian onto the normal's tangent plane to form the shape operator and extracting its eigenvectors.
We obtain the principal directions of curvature by projecting the Hessian onto the tangent plane and extracting its eigenvectors.
(v)~\emph{GSplats:} after the training we treat all splats as a point cloud and filter out those that have high depth error. We extract the features using the same protocol as in (ii). See \Cref{fig:teaser}. 

\vspace{-1mm}
\paragraph{Accuracy and Convergence on Spheres.}
We begin on the unit sphere, where closed-form solutions for heat diffusion and Poisson problems are available via spherical harmonics, enabling precise accuracy and convergence studies (see Chapter 6 \cite{atkinson2012spherical}). We also compare across four sphere mesh resolutions—\emph{coarse}, \emph{medium}, \emph{fine}, and \emph{very fine} with approximately $0.1k,\;1k,\;10k,\;100k$ vertices, respectively. 

Across resolutions, our solver matches CPM in accuracy (\Cref{tab:sphere_poisson_comparison}), even when CPM benefits from dense meshes, and follows similar error trends. Experiments further show that high-resolution SFEM provides a reliable proxy for ground truth (used later when analytic solutions are unavailable). Unlike mesh-centric pipelines, our errors are notably stable under remeshing and connectivity changes (see supplemental), indicating reduced sensitivity to sampling irregularities and local topology. Most importantly, our method operates natively in the neural implicit domain and can be used as a drop-in \emph{neural PDE layer} within standard deep-learning frameworks.
\camready{CPM and our method share the same band-based PDE solvers \camchange{and the runtime records both the extension step and the band solve step, i.e., the full pipeline.}{; runtimes include the full pipeline (extension + band solve).}}

\begin{table}[h]
    \centering
    \caption{
    \textbf{Poisson on the sphere (analytic GT).} Error vs.\ resolution for SFEM, CPM, and our method. We report normalized mean (NMAE) and max (NMaxE) errors (lower is better); all methods use identical right–hand sides and evaluation grids. \camready{Runtimes include the full pipeline; CPM and our method share the same band-based PDE solvers, isolating differences to the extension step.} See the supplemental for the corresponding heat equation table.}
    \label{tab:sphere_poisson_comparison}
    \footnotesize
    \begin{tabular}{rcccc}
        \toprule
        \textbf{Solver} & \textbf{Resolution} & \(\textbf{NMAE}\) $\downarrow$ & \(\mathbf{NMaxE}\) $\downarrow$ & \textbf{Time (s)}\\
        \midrule
        \multirow{3}{*}{SFEM} 
            & Coarse & $1.05 \times 10^{-2}$ & $2.32 \times 10^{-2}$ & 0.06 \\
            & Medium & $6.48 \times 10^{-4}$ & $1.90 \times 10^{-3}$ & 0.08 \\
            & Fine   & $2.84 \times 10^{-4}$ & $6.30 \times 10^{-4}$ & 0.23 \\
            & Very fine   & $1.11 \times 10^{-4}$ & $1.29 \times 10^{-4}$ & 5.46 \\
        \midrule
        % \multirow{3}{*}{CPM} 
            % & Coarse & $7.51 \times 10^{-2}$ & $5.58 \times 10^{-1}$ & 10.6 \\
            % & Medium & $1.80 \times 10^{-2}$ & $4.30 \times 10^{-2}$ & 22.3 \\
            % & Fine   & $1.68 \times 10^{-2}$ & $4.10 \times 10^{-2}$ & 24.2 \\
        % \midrule
        \multirow{3}{*}{CPM} 
            & Coarse & $4.56 \times 10^{-2}$ & $1.36 \times 10^{-1}$ & 10.5 \\
            & Medium & $1.49 \times 10^{-2}$ & $3.59 \times 10^{-2}$ & 24.4 \\
            & Fine   & $1.46 \times 10^{-2}$ & $3.49 \times 10^{-2}$ & 30.2 \\
            & Very fine & $1.48 \times 10^{-2}$ & $3.52 \times 10^{-2}$ & 335.0 \\
        \midrule
        \multirow{3}{*}{\textbf{Ours}} 
            & Coarse & $2.75 \times 10^{-2}$ & $9.14 \times 10^{-1}$ & 5.10 \\
            %& Medium & $1.34 \times 10^{-2}$ & $3.22 \times 10^{-2}$ \\%  & 25.5 \\
            & Medium & $1.24 \times 10^{-2}$ & $2.99 \times 10^{-2}$ & 25.5 \\
            & Fine   & $1.33 \times 10^{-2}$ & $3.17 \times 10^{-2}$ & 38.0 \\
            & Very fine & $1.32 \times 10^{-2}$ & $3.23 \times 10^{-2}$ & 72.6 \\
        \bottomrule
    \end{tabular}
\end{table}

\paragraph{Generalization on ShapeNet shapes.}
\camready{We evaluate the Poisson equation on complex shapes from \textbf{ShapeNet}, using a model trained only on \textit{spike} (\textit{spike-only}). We compare against CPM (training-free) and GINO. SFEM serves as reference, and we report NRMSE (scaled by $10^{-2}$). Results on five unseen shapes (A–E; chairs and tables) are shown in \Cref{tab:shapenet_generalisation}. Despite GINO yielding the best performance after overfitting on these shapes, we consistently outperform GINO on the same train-test split, indicating that our method generalizes to new data. This generalization also extends to unseen input functions for heat diffusion (see \Cref{fig:heat_table}). Additional results are provided in the supplemental \ref{sec:high_freq}.}
\begin{table}[h]
\centering
\footnotesize
\setlength{\tabcolsep}{4pt}
\caption{\camready{\textbf{Poisson equation on ShapeNet.} NRMSE ($\times 10^{-2}$) for five unseen shapes (A–E; chairs, tables); SFEM  as a reference.}}
\begin{tabular}{rccccc}
\toprule
\textbf{Method} & \textbf{A} & \textbf{B} & \textbf{C} & \textbf{D} & \textbf{E} \\
\midrule
% GINO (train) & 0.33 & 0.21 & 0.34 & 0.16 & 0.18 \\
GINO (spike-only) & 13.59 & 13.89 & 8.58 & 6.67 & 7.38 \\
CPM & 1.92 & 0.892 & 0.936 & 0.889 & 0.600\\
\textbf{Ours} (spike-only) & \textbf{1.05} & \textbf{0.909} & \textbf{0.986} & \textbf{0.354} & \textbf{0.364}\\
\midrule
% GINO (More shapes) & 4.21 & 11.29 & 3.00 & 8.63 & 2.95 \\
GINO (Overfitting) & 0.33 & 0.21 & 0.34 & 0.16 & 0.18 \\
\bottomrule
\end{tabular}
\label{tab:shapenet_generalisation}
\end{table}

\begin{figure}[h]
    \centering
    \includegraphics[width=.9\linewidth]{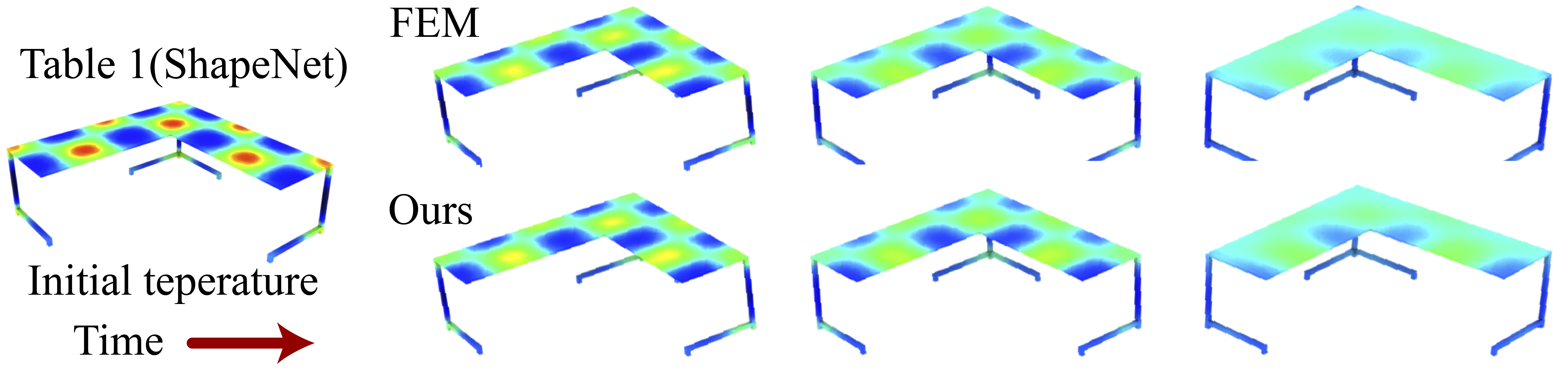}
    \caption{\camready{\textbf{Generalization to unseen input functions.} Heat diffusion over time for a high-frequency initial condition. Top: FEM reference. Bottom: Ours. Despite being trained on polynomial signals, our method accurately captures the dynamics, demonstrating robustness beyond the training distribution.}}
    \label{fig:heat_table}
\end{figure}

\begin{figure}[h!]
    \centering
    \includegraphics[width=\columnwidth]{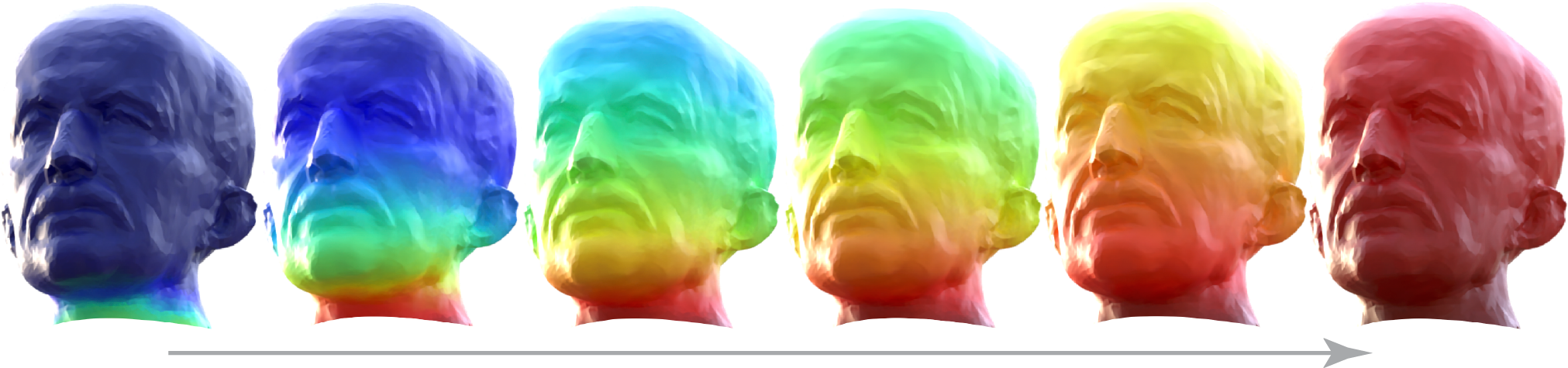}
    \caption{
    \textbf{Dirichlet boundaries on an open surface.}
    Heat diffusion on the Max Planck head cut at the neck (left to right) with boundary values clamped on the cut. See supplemental.}
    \label{fig:boundaryMax}
\end{figure}

\paragraph{Boundary handling.}
As in CPM, Neumann conditions are naturally satisfied since our update enforces normal consistency. For exterior Dirichlet boundaries, we follow the CPM practice of clamping boundary values and updating only in the band. On the Max Planck head (see \cref{fig:boundaryMax}), we solve heat with homogeneous and sinusoidal Dirichlet data and compare to a high-resolution SFEM reference on the corresponding open surface. Errors remain low and stable; detailed plots and per-case statistics are provided in the supplemental. These results confirm that exterior Dirichlet conditions are handled effectively, leveraging CPM’s boundary treatment, which fits naturally within our method.

\paragraph{Differentiable Application (Toy Example).}
\camready{
We demonstrate end-to-end differentiability by optimizing a scalar heat source intensity $h$ through our PDE solver. Given a target temperature profile corresponding to $h^\star=1$, }
\begin{wrapfigure}{r}{0.5\linewidth}
  \centering
  %\vspace{1pt}
  \hspace{-50pt}
  \includegraphics[width=\linewidth]{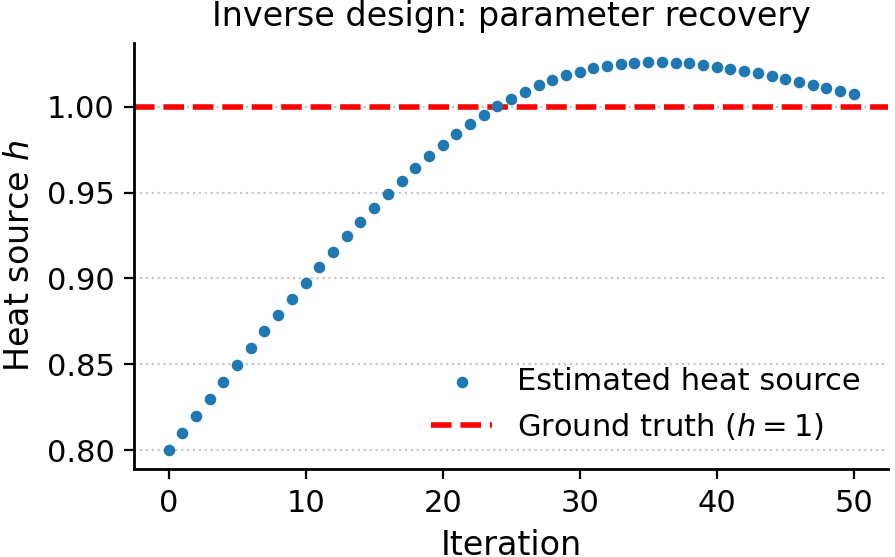}
  \hspace{-30pt}
  \vspace{-5pt}
  \label{fig:inv_problem}
\end{wrapfigure}
\camready{we initialize from \mbox{$h_0=0.8$} and minimize an $L^2$ loss by backpropagating through the full solver. As shown in the inset figure, $h$ converges to $1.0002$, confirming accurate gradient propagation.}

\paragraph{Ablations.}
Unless noted, all ablations use identical data, schedules, and hyperparameters; results across tables are not directly comparable. Overall, the studies support a simple, data-efficient design (see supplemental).
\emph{(i) Geometric consistency vs.\ data.}
A small weight on ${L}_{\mathrm{NC}}$ reliably lowers errors; large weights bias toward trivial normal-invariant fields and hurt fidelity. Even without ${L}_{\mathrm{NC}}$, the operator largely maintains normal consistency.
\emph{(ii) Local features.}
Positions and normals matter most; accurate curvature adds marginal benefit.
\emph{(iii) Band receptive field.}
Larger $k$ yields diminishing returns: errors drop mildly then plateau, consistent with convex aggregation. A mid-range $k$ ($\sim\!400$) balances accuracy, stability, and cost.
\emph{(iv) Model capacity.}
Shallow, narrow MLPs suffice; deeper/wider variants bring marginal gains.
\emph{(v) Learnable attention strength.}
A single learnable scalar $\lambda$ modestly but consistently improves accuracy, providing lightweight adaptation.

%\paragraph{Summary.}
%We benchmarked accuracy and runtime against SFEM/CPM, studied sensitivity to band size and mesh variation, compared across input modalities, and ablated key design choices. Additional plots (convergence curves, qualitative fields, overlap/aggregation studies) appear in the supplemental.

% !TEX root = ../main.tex

\section{Conclusion}
\camready{We presented a mesh-free, differentiable solver that operates directly on both neural and classical surface representations. By learning the extension step of embedding methods, we enable a unified, representation-agnostic approach to solving surface PDEs, which generalizes across shapes, topologies, and input functions without requiring per-instance optimization.}
%We presented a mesh-free solver for surface PDEs that operates \emph{directly} on neural and classical surface representations. Our geometry-conditioned local operator performs a direct  grid-to-grid update in a \emph{narrow band} around the target surface, remains fully differentiable, and generalizes across shapes, topologies and input functions without requiring meshing or per-instance optimization. 
\camready{On analytic benchmarks and across diverse surface representations, our method achieves competitive accuracy and runtime while avoiding mesh extraction and explicit extension steps, enabling a unified treatment across modalities.}
% On analytic benchmarks and real neural assets (across different representations), our method achieves competitive accuracy and runtime while eliminating mesh extraction and extend–restrict shuttling as required in the classical CPM method. 
\camready{We show that learning the extension step of embedding methods is sufficient to turn classical PDE solvers into representation-agnostic, differentiable operators.}
The approach integrates naturally into learning pipelines, \camchange{allowing us}{opening up options} to directly add PDE neural layers for analysis, editing, and reconstruction involving neural representations. While our focus is on the \emph{geometric} component of PDE solving—rather than optimizing high-order numerical schemes—this design makes our operator complementary to stronger discretizations and suitable for integration into more advanced solvers. This opens promising directions for coupling our geometry-aware update with advanced solvers, enabling PDE layers operating directly on neural surfaces while remaining compatible with classical surface representations.

\subsubsection*{Limitations and Future Work}

\emph{Self-intersections and medial-axis neighborhoods.}
\camready{Near self-intersections or the medial axis, SDF gradients can become unreliable, degrading update quality. This limitation is inherent to closest-point-based approaches. A potential remedy is a hybrid fallback, where ill-conditioned patches are detected (e.g., via $|\nabla \phi|$ or curvature thresholds) and handled with classical solvers. Our approach also relies on the underlying surface representation being accurate near the surface.}

\noindent
\emph{Evolving surfaces and rebanding.}
When the underlying surface moves under the PDE (e.g., surface evolution under curvature flow), the \emph{narrow band} must be rebuilt and re-sampled, reducing any amortization benefits. %Incremental band updates, and reusing cached features can mitigate cost; 
Extending the operator to predict both updates and band maintenance is also an interesting future direction.

\noindent
\emph{Grid dependence and scale effects.}
Our operator is not fully discretization-invariant: performance can vary with grid spacing and band thickness. Scale-aware conditioning and multi-resolution training may improve robustness. Learning on a modest range of resolutions generalizes in practice. Extending our coverage condition, it will be interesting to derive a relation between the sampling density and implicit surface quality.

%\noindent
%\emph{Scope of PDEs.}
%\camready{We focus on Poisson and heat equations; more challenging settings (e.g., stiff or anisotropic systems) may require tailored stabilization or implicit schemes. Extending our framework with learnable preconditioners or implicit updates is a promising avenue.}
%
% Finally, in this work, we focus on Poisson solves and heat equation; strongly anisotropic or stiff systems may require tailored stabilization or implicit time-stepping. Incorporating learnable preconditioners, implicit updates, or operator splitting within our framework is a promising future work.

\paragraph{Acknowledgments.} 
\textit{
We thank \href{https://hxwork.github.io/}{Hao Xu}, \href{https://people.mpi-inf.mpg.de/~nkairand/}{Navami Kairanda},                    \href{https://nathandking.github.io/}{Nathan King},  \href{https://github.com/RemySabathier}{Rémy Sabathier} and \href{https://romyjw.github.io/main/home.html}{Romy Williamson} for their feedback during the research. }

{
\small
\bibliographystyle{ieeenat_fullname}
\bibliography{neuralClosestPointMethod}
}

%\newpage
\clearpage
\setcounter{page}{1}
\setcounter{section}{0}
\setcounter{table}{0}
\maketitlesupplementary

\section{Ablation and Design Analysis.}
Each ablation isolates a single component of our method: we modify one factor at a time while keeping all other settings strictly identical (same dataset splits, schedules, and hyperparameters). This ensures that the observed variations can be attributed to the component being modified. Because ablations are run separately, with potentially different model variants, the numerical values reported across different tables are not meant to be compared to each other, only within each table. \textit{Our main empirical findings are:}
\begin{itemize}
    \item Normal-consistency term improves stability; $\alpha=10^{-2}$ performs best, though $\mathrm{MSE}$-only already yields good results. 
    \item Positions and normals are the essential surface features.
    \item A local band size of $k\approx400$ offers the best accuracy–runtime trade-off.
    \item Very shallow MLPs (2 layers) are sufficient.
    \item A learnable $\lambda$ improves accuracy.
    \item Patch overlap has negligible effect; local behavior dominates.
\end{itemize}

\subsection{Effect of the Normal-Consistency Loss}
We first assess the impact of the normal-consistency term $L_{\mathrm{NC}}$, which enforces invariance of the predicted function along surface normals. From Table~\ref{tab:ablation_losses_compact}, we observe that adding the normal-consistency term $L_{\mathrm{NC}}$ consistently improves the accuracy of the predicted solutions on both the analytical sphere and the Star Fruit surface (evaluated against SFEM as ground truth). In particular, the configuration $L = L_{\mathrm{MSE}} + 10^{-2}\,L_{\mathrm{NC}}$ yields the best average performance, achieving the lowest NMAE and NMaxE on both benchmarks. While the $L_{\mathrm{MSE}}$-only setting also produces low errors, it does not explicitly enforce normal invariance, and the model must learn this behavior implicitly from the data. Adding a small $L_{\mathrm{NC}}$ term reliably improves consistency and reduces error. Therefore, we retain the mixed formulation with coefficient $10^{-2}$ as our default choice for all subsequent experiments.

\begin{table}[h!]
    \centering
    \caption{Ablation on loss function design. 
    GT is analytical for the Sphere and SFEM for Star Fruit surfaces.}
    \label{tab:ablation_losses_compact}

    % --- Sphere ---
    \begin{tabular}{lcc}
        \toprule
        \multicolumn{3}{c}{\textbf{Sphere (GT analytic)}} \\
        \midrule
        \textbf{Loss} & NMAE $\downarrow$ & NMaxE $\downarrow$ \\
        \midrule
        $L_{\text{MSE}}$ & $2.29 \times 10^{-2}$ & $\underline{5.71 \times 10^{-2}}$ \\
        $+\,10^{-4}\,L_{\text{NC}}$ & $\underline{2.25 \times 10^{-2}}$ & $5.74 \times 10^{-2}$ \\
        $+\,10^{-3}\,L_{\text{NC}}$ & $2.31 \times 10^{-2}$ & $5.88 \times 10^{-2}$ \\
        $+\,10^{-2}\,L_{\text{NC}}$ & $\mathbf{2.08 \times 10^{-2}}$ & $\mathbf{5.43 \times 10^{-2}}$ \\
        $+\,10^{-1}\,L_{\text{NC}}$ & $2.33 \times 10^{-2}$ & $5.86 \times 10^{-2}$ \\
        $+\,L_{\text{NC}}$ & $2.40 \times 10^{-2}$ & $5.98 \times 10^{-2}$ \\
        \bottomrule
    \end{tabular}

    \vspace{0.4em}

    % --- General surfaces ---
    \begin{tabular}{lcc}
        \toprule
        \multicolumn{3}{c}{\textbf{Star Fruit (GT SFEM)}} \\
        \midrule
        \textbf{Loss} & NMAE $\downarrow$ & NMaxE $\downarrow$ \\
        \midrule
        $L_{\text{MSE}}$ & $1.54 \times 10^{-2}$ & $4.03 \times 10^{-2}$ \\
        $+\,10^{-4}\,L_{\text{NC}}$ & $1.51 \times 10^{-2}$ & $3.99 \times 10^{-2}$ \\
        $+\,10^{-3}\,L_{\text{NC}}$ & $1.57 \times 10^{-2}$ & $4.10 \times 10^{-2}$ \\
        $+\,10^{-2}\,L_{\text{NC}}$ & $\mathbf{1.31 \times 10^{-2}}$ & $\mathbf{3.56 \times 10^{-2}}$ \\
        $+\,10^{-1}\,L_{\text{NC}}$ & $\underline{1.34 \times 10^{-2}}$ & $\underline{3.63 \times 10^{-2}}$ \\
        $+\,L_{\text{NC}}$ & $1.42 \times 10^{-2}$ & $3.85 \times 10^{-2}$ \\
        \bottomrule
    \end{tabular}
\end{table}

\subsection{Surface Features Ablation} 
\label{sec:sf_ablations}
Before selecting a final feature set, we evaluate how different geometric cues affect performance. Our architecture is feature-agnostic: in principle, any per-surface descriptor (positional, differential, learned, or otherwise) can be injected into the operator. Modulo training, the operator can adapt to whichever features it receives. To quantify the importance of explicit geometric cues, we remove subsets of surface features (e.g., normals, curvature) and report the resulting degradation in performance. As shown in Table~\ref{tab:ablation_surface_features}, removing positional coordinates or surface normals degrades performance, confirming their importance for capturing local geometry and spatial context. In contrast, mean and Gaussian curvatures have little to no effect and may even introduce noise, slightly reducing accuracy. This suggests that curvature signals are either redundant with the local spatial and normal information, or too noisy to provide additional benefits. Interestingly, the performance drop remains moderate even when most surface features are removed: since the local band is constructed around the surface, it already provides rich geometric context and implicitly encodes high-level information about the shape, which helps maintain reasonable accuracy. In the final model, we therefore retain only positions and normals as surface features.

\begin{table}[h!]
    \centering
    \caption{Ablation on surface features. GT is analytical for the Sphere and SFEM-based for the Star Fruit surface.}
    \label{tab:ablation_surface_features}

    % --- Sphere ---
    \begin{tabular}{lcc}
        \toprule
        \multicolumn{3}{c}{\textbf{Sphere (GT analytic)}} \\
        \midrule
        \textbf{Features} & NMAE $\downarrow$ & NMaxE $\downarrow$ \\
        \midrule
        All features        & $\mathbf{1.94 \times 10^{-2}}$ & $\underline{5.20 \times 10^{-2}}$ \\
        w/o normals      & $2.10 \times 10^{-2}$ & $5.42 \times 10^{-2}$ \\
        w/o curvatures   & $\underline{1.98 \times 10^{-2}}$ & $\mathbf{5.19 \times 10^{-2}}$ \\
        w/o points       & $2.22 \times 10^{-2}$ & $5.62 \times 10^{-2}$ \\
        Points only         & $2.15 \times 10^{-2}$ & $5.44 \times 10^{-2}$ \\
        % w/o everything   & $2.16 \times 10^{-2}$ & $5.58 \times 10^{-2}$ \\
        \bottomrule
    \end{tabular}

    \vspace{0.4em}

    % --- Star Fruit ---
    \begin{tabular}{lcc}
        \toprule
        \multicolumn{3}{c}{\textbf{Star Fruit (GT SFEM)}} \\
        \midrule
        \textbf{Features} & NMAE $\downarrow$ & NMaxE $\downarrow$ \\
        \midrule
        All features        & $\underline{1.15 \times 10^{-2}}$ & $\underline{3.24 \times 10^{-2}}$ \\
        w/o normals      & $1.33 \times 10^{-2}$ & $3.64 \times 10^{-2}$ \\
        w/o curvatures   & $\mathbf{1.08 \times 10^{-2}}$ & $\mathbf{3.08 \times 10^{-2}}$ \\
        w/o points       & $1.25 \times 10^{-2}$ & $3.52 \times 10^{-2}$ \\
        Points only         & $1.36 \times 10^{-2}$ & $3.68 \times 10^{-2}$ \\
        % w/o everything   & $1.40 \times 10^{-2}$ & $3.77 \times 10^{-2}$ \\
        \bottomrule
    \end{tabular}
\end{table}

\subsection{Effect of the Local Band Size $k$}  
We vary the number of neighbouring band points $k$ per patch to analyse their impact on the performance. As shown in Table~\ref{tab:band_k}, increasing the local band size $k$ leads to systematically lower errors for both the analytical sphere and the Star Fruit surface. However, the improvement is marginal. This behavior can be explained by the architecture of our network: in the final aggregation stage, the model performs a convex combination of the band contributions, assigning smaller attention weights to points that are farther from the query. Consequently, distant points have little influence on the prediction, which limits the benefit of further enlarging the band. Choosing $k$ therefore involves a trade-off between accuracy, computational cost (since larger $k$ increases the number of parameters), and geometric flexibility: $k$ directly constrains the admissible band thickness and the spacing between band points, which are all coupled by a geometric inequality. In practice, we fix $k=400$, which provides a good balance between precision, stability, and efficiency.

\begin{table}[h!]
\centering
\caption{Effect of local band size $k$ on Poisson. Sphere uses analytical GT; Star Fruit uses SFEM GT.}
\label{tab:band_k}
\begin{tabular}{lcc}
\toprule
\textbf{Sphere (GT analytic)} & NMAE $\downarrow$ & NMaxE $\downarrow$ \\
\midrule
$k=25$  & $3.51 \times 10^{-2}$ & $8.85 \times 10^{-2}$ \\
$k=50$  & $3.47 \times 10^{-2}$ & $8.68 \times 10^{-2}$ \\
$k=100$ & $3.46 \times 10^{-2}$ & $8.67 \times 10^{-2}$ \\
$k=150$ & $3.38 \times 10^{-2}$ & $8.58 \times 10^{-2}$ \\
$k=200$ & $3.13 \times 10^{-2}$ & $8.09 \times 10^{-2}$ \\
$k=250$  & $3.10 \times 10^{-2}$ & $7.92 \times 10^{-2}$ \\
$k=300$  & $3.10 \times 10^{-2}$ & $7.92 \times 10^{-2}$ \\
$k=350$  & $3.05 \times 10^{-2}$ & $7.84 \times 10^{-2}$ \\
$k=400$  & $2.96 \times 10^{-2}$ & $7.70 \times 10^{-2}$ \\
$k=450$  & $2.85 \times 10^{-2}$ & $7.49 \times 10^{-2}$ \\
$k=500$  & $2.83 \times 10^{-2}$ & $7.33 \times 10^{-2}$ \\
\bottomrule
\end{tabular}

\vspace{0.4em}

\begin{tabular}{lcc}
\toprule
\textbf{Star Fruit (GT SFEM)} & NMAE $\downarrow$ & NMaxE $\downarrow$ \\
\midrule
$k=25$  & $2.94 \times 10^{-2}$ & $7.38 \times 10^{-2}$ \\
$k=50$  & $2.91 \times 10^{-2}$ & $7.34 \times 10^{-2}$ \\
$k=100$ & $2.88 \times 10^{-2}$ & $7.31 \times 10^{-2}$ \\
$k=150$ & $2.77 \times 10^{-2}$ & $7.12 \times 10^{-2}$ \\
$k=200$ & $2.36 \times 10^{-2}$ & $6.21 \times 10^{-2}$ \\
$k=250$ & $2.24 \times 10^{-2}$ & $5.95 \times 10^{-2}$ \\
$k=300$ & $2.14 \times 10^{-2}$ & $5.73 \times 10^{-2}$ \\
$k=350$ & $2.13 \times 10^{-2}$ & $5.68 \times 10^{-2}$ \\
$k=400$ & $1.93 \times 10^{-2}$ & $5.26 \times 10^{-2}$ \\
$k=450$ & $1.75 \times 10^{-2}$ & $5.04 \times 10^{-2}$ \\
$k=500$ & $1.53 \times 10^{-2}$ & $4.58 \times 10^{-2}$ \\
\bottomrule
\end{tabular}
\end{table}
% \begin{figure} \centering \includegraphics[width=1\linewidth]{Images/band_size_vs_error.pdf} \caption{Effect of the local band size $k$ on the accuracy of the Poisson equation.} \label{fig:band_size_vs_error} \end{figure}

\subsection{Network Architecture Ablation} 
We evaluate the influence of our MLP design by ablating the number of layers and the hidden width. We also tested standard nonlinearities (ReLU, SiLU, GELU) and observed negligible differences, so we keep ReLU for all experiments for consistency. As shown in Table~\ref{tab:ablation_query_mlp}, increasing the depth or width of the Query and Local Band MLPs beyond the configuration 2 layers and 64 neurons does not lead to significant improvement. The best trade-off between accuracy and efficiency is achieved with the simplest setting (2 layers, 64 neurons), which already provides sufficient expressivity for both the analytical Sphere and the Star Fruit surface. Deeper or wider architectures slightly overfit and do not generalize better, suggesting that the geometric encoding remains well captured by compact networks.
\begin{table}[h!]
    \centering
    \caption{Ablation on the Query and Local Band MLP architectures (layers $\times$ width). Both MLPs share the same structure since their output dimensions must match. Each cell reports NMAE / NMaxE, with all values scaled by $10^{-2}$. GT is analytical for the Sphere and SFEM-based for the Star Fruit surface.}
    \label{tab:ablation_query_mlp}

    % --- Sphere ---
    \begin{tabular}{lccc}
        \toprule
        \multicolumn{4}{c}{\textbf{Sphere (GT analytic)}} \\
        \midrule
        \diagbox{Layers}{Width} & 32 & 64 & 128 \\
        \midrule
        2 & 2.23 / 5.63 & \textbf{2.08 / 5.43} & 2.19 / 5.61 \\
        3 & 2.27 / 5.69 & 2.24 / 5.68 & 2.24 / 5.70 \\
        4 & 2.32 / 5.83 & \underline{2.18 / 5.45} & 2.24 / 5.77 \\
        \bottomrule
    \end{tabular}

    \vspace{0.4em}

    % --- Star Fruit ---
    \begin{tabular}{lccc}
        \toprule
        \multicolumn{4}{c}{\textbf{Star Fruit (GT SFEM)}} \\
        \midrule
        \diagbox{Layers}{Width} & 32 & 64 & 128 \\
        \midrule
        2 & \underline{1.36 / 3.67} & \textbf{1.31 / 3.56} & 1.43 / 3.82 \\
        3 & 1.48 / 3.92 & 1.43 / 3.84 & 1.41 / 3.81 \\
        4 & 1.55 / 4.07 & \underline{1.36 / 3.67} & 1.45 / 3.92 \\
        \bottomrule
    \end{tabular}
\end{table}

As shown in Table~\ref{tab:ablation_surface_mlp}, increasing the depth or width of the Surface Features MLP brings only marginal improvements. The overall trend is consistent across both the analytical Sphere and the Star Fruit surface: moderate configurations (2–3 layers with 64 neurons) already achieve near-optimal performance. This indicates that the local geometric features are relatively low-dimensional and can be effectively captured by shallow networks. Deeper or wider models yield no significant benefit, confirming that compact architectures provide the best balance between expressivity, generalization, and computational cost.

\begin{table}[h!]
    \centering
    \caption{Ablation on the surface features MLP architecture (layers $\times$ width). Each cell reports NMAE / NMaxE, with all values scaled by $10^{-2}$. GT is analytical for the Sphere and SFEM-based for the Star Fruit surface.}
    \label{tab:ablation_surface_mlp}

    % --- Sphere ---
    \begin{tabular}{lccc}
        \toprule
        \multicolumn{4}{c}{\textbf{Sphere (GT analytic)}} \\
        \midrule
        \diagbox{Layers}{Width} & 32 & 64 & 128 \\
        \midrule
        2 & 2.23 / 5.63 & \textbf{2.08 / 5.43} & 2.19 / 5.61 \\
        3 & 2.27 / 5.69 & 2.24 / 5.68 & 2.24 / 5.70 \\
        4 & 2.32 / 5.83 & \underline{2.18 / 5.45} & 2.24 / 5.77 \\
        \bottomrule
    \end{tabular}

    \vspace{0.4em}

    % --- Star Fruit ---
    \begin{tabular}{lccc}
        \toprule
        \multicolumn{4}{c}{\textbf{Star Fruit (GT SFEM)}} \\
        \midrule
        \diagbox{Layers}{Width} & 32 & 64 & 128 \\
        \midrule
        2 & 1.48 / 3.91 & \underline{1.31} / \textbf{3.56} & \underline{1.31 / 3.61} \\
        3 & 1.42 / 3.82 & 1.45 / 3.85 & 1.41 / 3.77 \\
        4 & \textbf{1.30} / 3.64 & 1.35 / 3.67 & 1.36 / 3.65 \\
        \bottomrule
    \end{tabular}
\end{table}

\subsection{Influence of the Learnable Parameter $\lambda$}  
We study the role of the learnable scalar $\lambda$ that modulates the surface-aware penalty within each patch. As shown in Table~\ref{tab:influence_lambda}, introducing the learnable parameter $\lambda$ slightly improves accuracy on both analytical and SFEM-based benchmarks. This indicates that allowing the network to adapt the strength of the surface-aware penalty provides additional flexibility during training. Keeping $\lambda$ learnable is therefore beneficial and theoretically consistent: if this term were unnecessary, the optimization would naturally drive $\lambda$ toward $\approx1$.
\begin{table}[h!]
    \centering
    \caption{Ablation on the learnable $\lambda$ parameter in the attention formulation. 
    GT is analytical for the Sphere and SFEM-based for the Star Fruit surface.}
    \label{tab:influence_lambda}

    % --- Sphere ---
    \begin{tabular}{lcc}
        \toprule
        \multicolumn{3}{c}{\textbf{Sphere (GT analytic)}} \\
        \midrule
        \textbf{Configuration} & NMAE $\downarrow$ & NMaxE $\downarrow$ \\
        \midrule
        w/ learnable $\lambda$ & $\mathbf{2.01 \times 10^{-2}}$ & $\mathbf{5.28 \times 10^{-2}}$ \\
        w/o $\lambda$ (fixed to $1$) & $\underline{2.16 \times 10^{-2}}$ & $\underline{5.59 \times 10^{-2}}$ \\
        \bottomrule
    \end{tabular}

    \vspace{0.4em}

    % --- Star Fruit ---
    \begin{tabular}{lcc}
        \toprule
        \multicolumn{3}{c}{\textbf{Star Fruit (GT SFEM)}} \\
        \midrule
        \textbf{Configuration} & NMAE $\downarrow$ & NMaxE $\downarrow$ \\
        \midrule
        w/ learnable $\lambda$ & $\mathbf{1.10 \times 10^{-2}}$ & $\mathbf{3.15 \times 10^{-2}}$ \\
        w/o $\lambda$ (fixed to $1$) & $\underline{1.34 \times 10^{-2}}$ & $\underline{3.63 \times 10^{-2}}$ \\
        \bottomrule
    \end{tabular}
\end{table}

\subsection{Patch Overlap and Aggregation}
Finally, we explore the influence of patch overlap in the global reconstruction, varying the overlap ratio and the temperature parameter $T$ used in the convex aggregation weights. The temperature has only a minor effect on accuracy: by comparing the aggregated field against a ground-truth closest point extension, we found that a moderate value ($T \approx 0.5$) provides stable and smooth reconstructions. Since changes in $T$ had negligible numerical impact, we fix $T=0.5$ in all experiments. The results in Table \ref{tab:patches_importance} show that the error remains virtually unchanged as the number of patches increases—from 600 to 1250, both NMAE and NMaxE vary by less than $1\%$. This stability arises because the operator performs a local convex aggregation: the softmax weights strongly emphasize nearby band samples (because the parameter $T$ seen before is small), while assigning nearly zero weight to distant ones. As a consequence, enlarging the patch neighborhood adds points that contribute negligibly to the update. This saturation is consistent with our observations that the method relies primarily on local neighborhoods.
\begin{table}[h!]
    \centering
    \caption{
    Effect of patch overlap on Poisson (sphere, analytical GT). Increasing the number of patches has negligible impact on accuracy: because the operator performs a convex local aggregation with softmax weights centered around each query, distant samples contribute almost nothing.}
    \label{tab:patches_importance}

    \begin{tabular}{lcc}
        \toprule
        \textbf{Number of Patch} & \textbf{NMAE} $\downarrow$ & \textbf{NMaxE} $\downarrow$ \\
        \midrule
        $600$   & $1.33 \times 10^{-2}$ & $3.19 \times 10^{-2}$ \\
        $650$  & $1.32 \times 10^{-2}$ & $3.17 \times 10^{-2}$ \\
        $1250$   & $1.33 \times 10^{-2}$ & $3.18 \times 10^{-2}$ \\
        \bottomrule
    \end{tabular}
\end{table}

\subsection{Choice of Training Functions}
We explored several options for selecting a function family to supervise our operator:
\begin{itemize}
    \item \textbf{Sobolev spaces}: Sampling functions from Sobolev spaces would be theoretically natural, since PDE solutions belong to these spaces. However, Sobolev spaces do not offer an explicit basis, making them impractical for local supervision.
    \item \textbf{Laplacian eigenfunctions}: Eigenfunctions form an orthonormal basis of $L^2$, but they are tied to the global geometry of a specific surface (e.g., spherical harmonics on $\mathbb{S}^2$). Using them would bake in shape-dependent biases, whereas our goal is to generalize beyond the training geometry.
    \item \textbf{Monomials}: We therefore use low-degree monomials $(x,y,z) \longmapsto x^i y^j z^k$. PDE solutions are smooth, and our operator is local, so any smooth function can be well approximated by the first terms of its Taylor expansion. Monomials span exactly this local polynomial space and provide a simple, geometry-agnostic basis that works across shapes and modalities.
\end{itemize}
\section{Comparative Evaluation}

\subsection{Accuracy \& Convergence}
In this evaluation, we compare our method against classical solvers in terms of accuracy and runtime. Complementing the Poisson results in Table~\ref{tab:sphere_poisson_comparison}, Table~\ref{tab:sphere_heat_comparison} reports analogous results for the heat equation. SFEM benefits from decades of optimization in mesh-based numerical solvers, whereas our approach is still a research prototype. On this idealized benchmark, SFEM attains extremely low errors thanks to perfect mesh geometry and explicit discretization. When given the exact same geometric information (surface samples and normals extracted from the mesh), our method maintains stable accuracy across resolutions, with NRMSE consistently around $7\times 10^{-3}$ for the heat equation and NMAE around $1.3\times 10^{-2}$ for Poisson.
Although not as precise as SFEM on this setting, our operator is resolution-independent, requires no meshing, and relies on a single learned update rule that generalizes across PDEs, shapes and modalities. These results show that even when geometry comes from the mesh—rather than a neural surface—our solver behaves robustly and does not benefit from mesh refinement, unlike classical discretization-based methods. 
A single outlier appears in the NMaxE metric (on the coarse sphere in Table \ref{tab:sphere_poisson_comparison}), originating from a configuration that is ill-posed for our method. In this case, the solver receives almost no meaningful surface information: the local patch contains essentially a single isolated surface sample, which prevents the operator from inferring any reliable geometric structure. Unlike in the surface-features ablation (Sec~\ref{sec:sf_ablations} of the supplemental), where the narrow band is still well formed and the network can therefore produce good predictions even with limited features, here the band itself is poorly constructed, providing insufficient spatial context for the operator to recover a high-level representation of the underlying surface. Increasing the band thickness alleviates this issue: with a slightly larger band, the outlier NMaxE decreases from $9.14\times 10^{-1}$ to $5.26\times 10^{-1}$, while maintaining a good NMAE (around $4.2\times 10^{-2}$).
\begin{table}[h!]
    \centering
    \caption{
    \textbf{Heat equation on the sphere (analytic GT).} Error vs.\ resolution for SFEM, and our method. We report normalized root mean square (NRMSE) error (lower is better); all methods use identical initial condition and evaluation grids.}
    \label{tab:sphere_heat_comparison}

    \begin{tabular}{lcc}
        \toprule
        \textbf{Solver} & \textbf{Resolution} & \(\textbf{NRMSE}\) $\downarrow$ \\
        \midrule
        \multirow{4}{*}{SFEM} 
            & Coarse     & $2.93 \times 10^{-4}$  \\
            & Medium     & $1.50 \times 10^{-6}$   \\
            & Fine       & $1.58 \times 10^{-7}$  \\
        \midrule
        \multirow{4}{*}{\textbf{Ours}} 
            & Coarse     & $9.46\times10^{-3}$ \\
            & Medium     & $7.20\times10^{-3}$ \\
            & Fine       & $7.24\times10^{-3}$ \\
        \bottomrule
    \end{tabular}
\end{table}

Table~\ref{tab:sphere_relative_comparison} highlights a marked contrast in robustness to mesh resolution between FEM and our method. SFEM exhibits very large error variations when the mesh is coarsened—reaching increases of several orders of magnitude—while our operator remains remarkably stable, with variations typically within only a few percent. This confirms that accuracy is essentially resolution-independent for our approach. Naturally, FEM achieves higher absolute accuracy and lower runtime on this idealized sphere benchmark; its behavior here reflects decades of optimization and exact access to mesh-based geometry. Nonetheless, the comparison demonstrates that our method maintains consistent accuracy across discretizations, offering robustness where traditional solvers may degrade sharply under coarse or irregular meshing.

\begin{table}[h!]
    \centering
    \caption{
    Relative variation of accuracy and runtime for the Poisson equation on the sphere across mesh resolutions. For each solver, the \textit{Very fine} (VF) resolution is used as the reference (0\%) for accuracy, while the \textit{Coarse} (C) resolution serves as the reference (0\%) for runtime. Mesh resolutions follow the shorthand: C = Coarse, M = Medium, F = Fine, VF = Very Fine. Positive values denote degradation; negative values indicate improvement.}
    \label{tab:sphere_relative_comparison}
    \begin{tabular}{lcccc}
        \toprule
        \textbf{Solver} & \textbf{Res} & \(\mathbf{\Delta NMAE}\) & \(\mathbf{\Delta NMaxE}\) & $\mathbf{\Delta T}$ \\
        \midrule
        \multirow{3}{*}{SFEM} 
            & C & + 9359\% & + 17884\% & 0\% \\
            & M & + 483\% & + 1372\% & \textbf{+ 33\%} \\
            & F   & + 155\% & + 388\% & \textbf{+ 283\%} \\
            & VF   & 0 \% & 0\% & + 9000\% \\
        \midrule
        \multirow{3}{*}{\textbf{Ours}} 
            & C & \textbf{+ 97\%} & \textbf{+ 2686\%} & 0\% \\
            & M & \textbf{- 3\%} & \textbf{- 1\%} & + 400\% \\
            & F   & \textbf{- 1\%} & \textbf{- 1\%} & + 645\% \\
            & VF & 0\% & 0\% & \textbf{+ 1323\%} \\
        \bottomrule
    \end{tabular}
\end{table}

\subsection{Handling Different Shape Representations}
We run our solver on multiple shape encodings (mesh, point cloud, SNS, SDF/occupancy INRs, Gaussian Splatting), using the same local features described above (positions and normals, with optional curvature), illustrated in \Cref{fig:teaser}. Evaluation is performed on the steady-state solution, whose ground truth is analytic (given by the mean of the initial condition over the surface). Although a direct comparison is not strictly meaningful across different shapes and representations, similar trends are observed (see \Cref{tab:representation_comparison}), with SNS yielding the best results—consistent with its superior geometric estimates. Generalization across modalities also indicates robustness to noisy geometric quantities, as different representations provide features of varying quality. Overall, these results highlight that our method performs consistently well across diverse modalities, rather than aiming for direct performance comparison between them.

\begin{table}[h!]
    \centering
    \footnotesize
    \caption{
    Comparison of different \textbf{surface representations} for the heat equation on the steady state. The metric is the Normalized Root Mean Squared Error (NRMSE), computed against the analytic solution.}
    \label{tab:representation_comparison}

    \begin{tabular}{rc}
        \toprule
        \textbf{Representation / Shape} & \(\textbf{NRMSE}\) $\downarrow$ \\
        \midrule
        Neural SDF / Camera & $2.17\times10^{-2}$ \\
        Overfitted SDF / Max Planck Face & $3.03\times10^{-2}$ \\
        Spherical Neural Surface / Armadillo & $\mathbf{1.88\times10^{-2}}$ \\
        Gaussian Splatting / Snowman & $6.15\times10^{-2}$ \\
        Point Cloud / Hat & $2.94\times10^{-2}$ \\
        Mesh / Holey Human & $1.92\times10^{-2}$ \\
        \bottomrule
    \end{tabular}
\end{table}

\subsection{Generalization across shapes}
\begin{figure}[h!]
    \centering
    \includegraphics[width=\columnwidth]{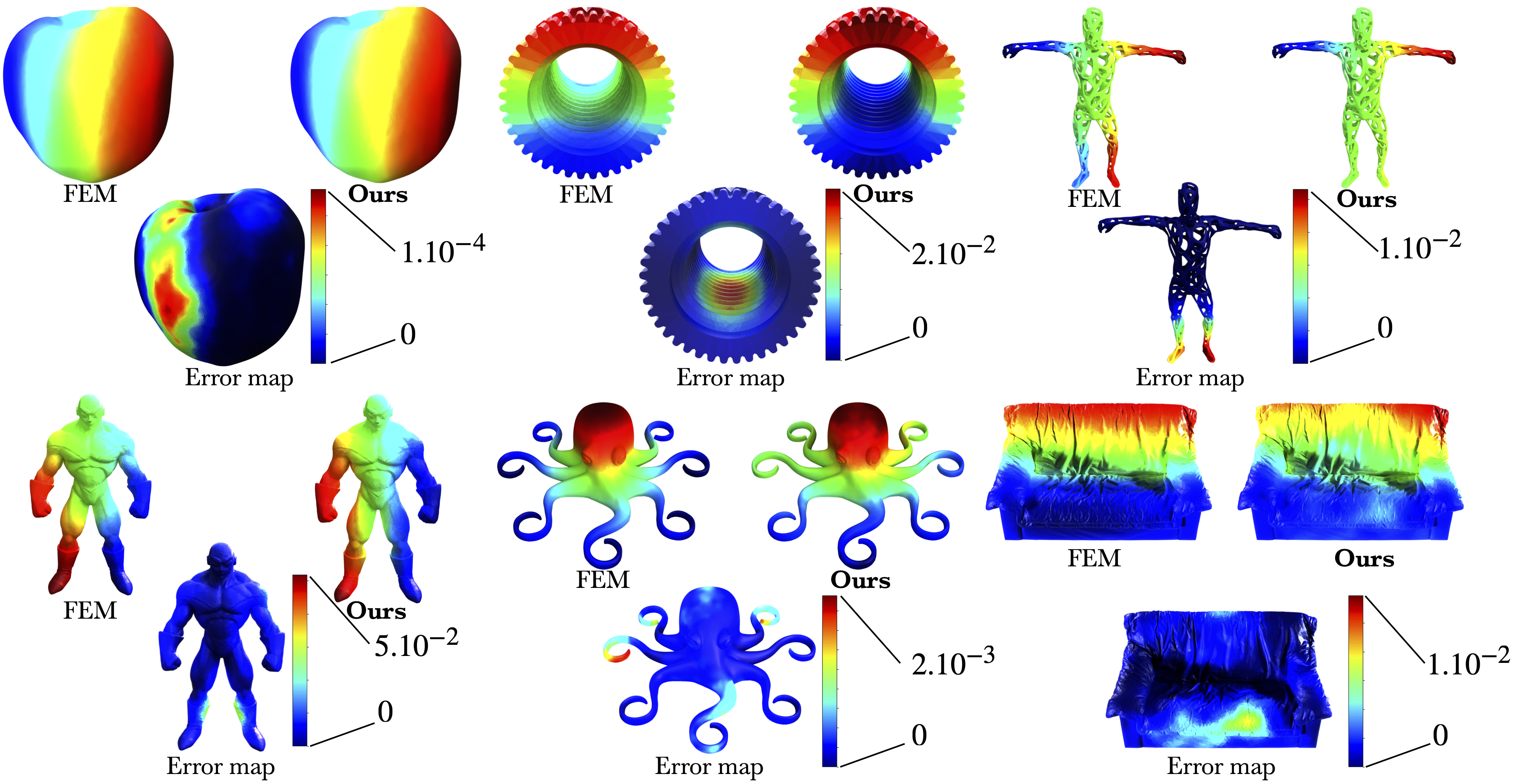}
    \caption{
    \textbf{Comparison to SFEM on diverse shapes.}
For each object, left shows SFEM and right shows ours; the small inset below visualizes the pointwise error (ours vs.\ SFEM) with a hot–cold colormap. 
See color bar for error scale and the supplemental for per-shape statistics. (Error colormaps are normalized per instance.) 
}
    \label{fig:results_gallery}
\end{figure}

Our neural operator, trained once on a single shape, transfers seamlessly to \emph{unseen} shapes and topologies across input modalities. As illustrated in \Cref{fig:results_gallery} and quantified in \Cref{tab:poisson_shapes}, it closely matches SFEM reference solutions for Poisson across diverse geometries (e.g., organic, CAD parts with sharp transitions, and thin-structure cases) with consistently low NRMSE. This amortized, geometry-conditioned behavior underpins cross-shape generalization. Nonetheless, errors tend to appear near regions where closest points are not unique (e.g., sharp edges, thin parts, near the medial axis), a limitation inherited from the CPM formulation.

\begin{table}[h!]
    \centering
    \footnotesize
    \caption{Poisson equation results on different shapes (NRMSE $\downarrow$). 
    Errors are computed against the FEM as ground truth solution.}
    \label{tab:poisson_shapes}
    \begin{tabular}{r c  r c}
        \toprule
        \textbf{Shape} & \textbf{NRMSE} & \textbf{Shape} & \textbf{NRMSE} \\
        \midrule
        Jared        & $5.12 \times 10^{-2}$ & Sofa         & $2.65 \times 10^{-2}$ \\
        Octopus      & $1.13 \times 10^{-2}$ & Holey Human  & $2.67 \times 10^{-2}$ \\
        Apple        & $3.20 \times 10^{-3}$ & Fastener      & $3.95 \times 10^{-2}$ \\
        \bottomrule
    \end{tabular}
    \vspace*{-.1in}
\end{table}
\subsection{Runtime scaling}
Table~\ref{tab:band_mesh_scaling} shows that the runtime of our method is primarily controlled by the size of the local band rather than by the resolution of the underlying surface. Increasing the number of band samples leads to an increase in cost, whereas changing the mesh from 1\text{k} to 100\text{k} vertices has only a low impact. Together with the robustness results in Table~\ref{tab:sphere_relative_comparison}, this confirms that our operator is effectively decoupled from the surface discretization and behaves consistently across meshes of different densities. We note that the reported runtimes include both the neural update and the PDE \emph{solve} step inside the band. The latter is independent of our contribution and could be further optimized using other standard numerical or hardware-specific accelerations.
\begin{table}[h!]
    \centering
    \caption{
    Computation time (in seconds) for the Poisson equation as a function of surface resolution (number of vertices) and band size $N$. The runtime scales almost only with the number of band samples, while remaining relatively insensitive to changes in mesh resolution. This confirms that our method’s complexity is primarily governed by the band size rather than the underlying surface discretization.}
    \label{tab:band_mesh_scaling}

    \begin{tabular}{lccc}
        \toprule
        \textbf{Band size} & \textbf{1k vertices} & \textbf{10k vertices} & \textbf{100k vertices} \\
        \midrule
        $N = 10k$   & 3.20s  & 4.13s & 11.8s  \\
        $N = 20k$  & 5.89s & 6.75s & 15.3s \\
        $N = 30k$  & 9.32s  & 10.7s & 20.2s \\
        $N = 40k$  & 16.4s & 18.2s & 27.6s \\
        $N = 50k$  & 30.1s & 35.6s & 67.1s\\
        \bottomrule
    \end{tabular}
\end{table}

\subsection{Robustness to noise}
\camready{
FEM is more accurate on high-quality fixed meshes; however, it is highly dependent on the fitting of the element mesh to the shape. In the table below, we show that, on the Jared model, the performance of FEM has higher variance with respect to the noise-free version when the input has different scales ($\sigma$) of noise.}

\begin{table}[h]
    \centering
    \footnotesize
    \caption{Error of added Gaussian noise to the mesh with different $\sigma$, compared to the FEM on the noise-free mesh}
    \begin{tabular}{lcc} 
    \toprule 
    \textbf{Method} & FEM & Ours\\ 
    \midrule $\boldsymbol{\sigma=0.0005}$  &  $ \mathbf{2.39\times 10^{-2}}$  & $ 3.16\times 10^{-2}$ \\
    $\boldsymbol{\sigma=0.001}$   &  $ \mathbf{3.02 \times 10^{-2} }$ & $ 3.18\times 10^{-2}$\\ 
    $\boldsymbol{\sigma=0.005}$   &  $ 7.06 \times 10^{-1} $ & $ \mathbf{3.17\times 10^{-2}}$ \\
    $\boldsymbol{\sigma=0.01}$    &  $ 8.64 \times 10^{-1} $ & $ \mathbf{3.18\times 10^{-2}}$ \\ 
    \bottomrule 
    \end{tabular}  
\label{tab:noise_comparsion}
\end{table}

\subsection{Robustness to remeshing}
To evaluate robustness to surface remeshing, we construct four triangulated discretizations of the unit sphere that share the same number of vertices ($N{=}1500$) but differing in their vertex sampling patterns:
\begin{itemize}
    \item Regular: uniform triangles with Delaunay.
    \item Random: points are drawn independently from an isotropic 3D Gaussian, then normalized onto the unit sphere
    \item Jittered: add a small Gaussian-sampled noise to evenly distributed samples and map to the unit sphere.
    \item Blue-noise: sampling points with a blue noise pattern (from a list of random points, choose the farthest to yield a well-spaced, random distribution).
\end{itemize}
Table~\ref{tab:sphere_mesh_variation} reports the Poisson error for SFEM and for our method, together with the relative variation across mesh types. SFEM is strongly affected by the sampling strategy: its error varies by almost an order of magnitude across the four meshes, reflecting its sensitivity to triangle quality and vertex distribution.
In contrast, our method shows negligible variation (within $\pm 3\%$ for NMAE and $\pm6\%$ for NMaxE), confirming that the learned operator is largely invariant to surface discretization.
This robustness stems from the fact that our solver operates in the narrow band with geometry extracted only locally, rather than relying on mesh-dependent differential operators.
\begin{table}[h!]
    \centering
    \caption{
    Robustness to mesh variation for the Poisson equation on the sphere. All meshes contain $N{=}1500$ vertices but differ in sampling strategy and triangulation. Errors are computed against the analytical solution. Top: absolute errors for both methods. Bottom: relative variation (min–max) w.r.t.\ the Regular mesh.}
    \label{tab:sphere_mesh_variation}

    % ---------- SFEM ----------
    \resizebox{0.95\linewidth}{!}{%
    \begin{tabular}{lcc}
        \toprule
        \multicolumn{3}{c}{\textbf{SFEM}} \\
        \midrule
        \textbf{Metric} & \textbf{Mesh type} & \textbf{Error} $\downarrow$ \\
        \midrule
        \multirow{4}{*}{NMAE}
            & Regular    & $6.48 \times 10^{-4}$ \\
            & Random     & $ 7.09 \times 10^{-3} \pm 1.85 \times 10^{-5} $ \\
            & Blue-noise & $ 2.18 \times 10^{-3} \pm 3.30 \times 10^{-6} $ \\
            & Jittered   & $ 1.12 \times 10^{-3} \pm 2.36 \times 10^{-9} $\\
        \cmidrule(lr){1-3}
        \multirow{4}{*}{NMaxE}
            & Regular    & $1.90 \times 10^{-3}$ \\
            & Random     & $ 1.03 \times 10^{-2} \pm 1.98 \times 10^{-5} $ \\
            & Blue-noise & $ 3.63 \times 10^{-3} \pm 3.88 \times 10^{-6} $ \\
            & Jittered   & $ 3.09 \times 10^{-3} \pm 9.18 \times 10^{-8} $ \\
        \bottomrule
    \end{tabular}}

    \vspace{-0.3em}

    % ---------- OURS ----------
    \resizebox{0.95\linewidth}{!}{%
    \begin{tabular}{lcc}
        \toprule
        \multicolumn{3}{c}{\textbf{Ours}} \\
        \midrule
        \textbf{Metric} & \textbf{Mesh type} & \textbf{Error} $\downarrow$ \\
        \midrule
        \multirow{4}{*}{NMAE}
            & Regular    & $1.34 \times 10^{-2}$ \\
            & Random     & $1.30 \times 10^{-2} \pm 4.24 \times 10^{-4}$ \\
            & Blue-noise & $1.39 \times 10^{-2} \pm 4.12 \times 10^{-4}$ \\
            & Jittered   & $1.37 \times 10^{-2} \pm 3.24 \times 10^{-4}$ \\
        \cmidrule(lr){1-3}
        \multirow{4}{*}{NMaxE}
            & Regular    & $3.22 \times 10^{-2}$ \\
            & Random     & $3.21 \times 10^{-2} \pm 2.60 \times 10^{-4}$ \\
            & Blue-noise & $3.34 \times 10^{-2} \pm 9.69 \times 10^{-4}$ \\
            & Jittered   & $3.42 \times 10^{-2} \pm 2.12 \times 10^{-3}$ \\
        \bottomrule
    \end{tabular}}

    % ---------- Comparaison % ----------
    \resizebox{0.95\linewidth}{!}{%
    \begin{tabular}{lcc}
        \toprule
        \multicolumn{3}{c}{\textbf{Relative variation}} \\
        \midrule
        \textbf{Method} & $\Delta$NMAE & $\Delta$NMaxE \\
        \midrule
        SFEM & +72\% to +994\% & +62\% to +442\% \\
        Ours & \textbf{-2\% to +3\%}  & \textbf{-0.3\% to +6\%} \\
        \bottomrule
    \end{tabular}}
\end{table}

\subsection{Generalization on high frequency functions}\label{sec:high_freq}
\camready{We evaluate the Poisson equation ($\Delta u = f_\omega$) with sinusoidal input of the form $f_{\omega}(\mathbf{x})=\sin(\omega x)\times\sin(\omega y)\times\sin(\omega z)$ of increasing frequency $\omega$ on a fixed shape. SFEM is used as reference, and we report NRMSE, see table \ref{tab:high_frequency}. 
\begin{table}[h]
\centering
\footnotesize
\caption{\camready{High-frequency stress test (NRMSE). Performance degrades with frequency, but our method remains stable and competitive.}}
\label{tab:high_frequency}
\setlength{\tabcolsep}{6pt}
\begin{tabular}{rccc}
\toprule
\textbf{Method} & $\boldsymbol{\omega=1}$ & $\boldsymbol{\omega=10}$ & $\boldsymbol{\omega=20}$ \\
\midrule
% GINO (train) & $3.33 \times 10^{-3}$ & $6.43 \times 10^{-3}$ & $1.03 \times 10^{-2}$ \\
% GINO (test) & $4.21 \times 10^{-2}$ & $2.62 \times 10^{-1}$ & $1.53 \times 10^{-1}$ \\
GINO (spike-only) & $1.36 \times 10^{-1}$ & $2.63 \times 10^{-1}$ & $4.21 \times 10^{-1}$ \\
CPM  & $1.92 \times 10^{-2}$ & $8.64 \times 10^{-2}$ & $5.34 \times 10^{-2}$ \\
Ours (spike-only) & $1.05 \times 10^{-2}$ & $3.05 \times 10^{-2}$ & $4.10 \times 10^{-2}$ \\
\midrule
GINO (overfitting) & $3.33 \times 10^{-3}$ & $6.43 \times 10^{-3}$ & $1.03 \times 10^{-2}$ \\
\bottomrule
\end{tabular}
\end{table}
Our method remains stable as frequency increases and consistently outperforms the spike-only GINO baseline, while achieving better performance than CPM. As expected, performance degrades for higher frequencies due to increased smoothness requirements, but the method retains good accuracy relative to the training regime.
We additionally show a qualitative heat example, with the sinusoidal pattern as the initial condition, with $\omega=10$, shown in figure \ref{fig:heat_table} (main article). }

\subsection{Dirichlet boundary condition}
% To generate controlled openings, we intersect the Max Planck model with horizontal planes at heights \(z\in\{-0.7,-0.6,-0.5,-0.4,-0.3\}\), ranging from an almost complete head (\(z=-0.7\)) to only the nose and above (\(z=-0.3\)). For each truncated surface, we solve the PDE with two types of Dirichlet boundary conditions on the cut curve $\partial\mathcal{S}$. 
To generate a controlled open boundary, we cut the Max Planck surface with a single horizontal plane located at the neck region. This creates a well-defined boundary curve $\partial\mathcal{S}$ on which we impose Dirichlet conditions. We test two boundary conditions of increasing difficulty: a simple, constant one, and a more oscillatory \emph{fancy} condition that stresses the solver’s ability to handle nontrivial boundary signals. We then solve the surface heat equation
\[
\frac{\partial u}{\partial t} = \Delta_{\mathcal{S}} u,
\qquad u(t=0)=f,
\]
on each truncated surface until reaching the steady state, which we compare against the FEM ground truth.

\begin{enumerate}
    \item \textbf{Constant Dirichlet condition}
    \[f = 0, \qquad u\big|_{\partial\mathcal{S}} = 1 .\]
    \item \textbf{Sinusoidal Dirichlet condition}
    \[
    \begin{aligned}
        f &= -\sin\!\big(k\,\arctan(x,z)\big), \\
        u\big|_{\partial\mathcal{S}} &= \sin\!\big(k\,\arctan(x,z)\big).
    \end{aligned}
    \]
\end{enumerate}
The results in Table~\ref{tab:heat_bc} show that our method remains stable under both boundary regimes, with only a moderate increase in error for the more oscillatory sinusoidal condition. This robustness is expected, since our solver inherits the boundary–condition handling of the original Closest Point Method~\cite{ruuth2008cpm}: Dirichlet values are directly imposed on the band nodes whose closest-point projections lie on the boundary curve $\partial \mathcal{S}$. As a result, the behavior at the boundary is preserved accurately even on complex geometries.
\begin{table}[h!]
    \centering
    \caption{Heat equation results on the truncated Max Planck surface. 
    Errors are computed against the FEM as the ground truth.}
    \label{tab:heat_bc}

    \begin{tabular}{lcc}
        \toprule
        \textbf{Boundary Conditions type} & \textbf{NRMSE} $\downarrow$ \\
        \midrule
        \emph{standard} & $2.85\times 10^{-2}$\\
        \emph{fancy}    & $3.40\times10^{-2}$  \\
        \bottomrule
    \end{tabular}
\end{table}
% \begin{table}[h!]\centering\caption{Error on truncated Max–Planck surfaces with exterior homogeneous Dirichlet boundary conditions.}\label{tab:boundary}\begin{tabular}{lcc}\toprule\textbf{Level} & \textbf{NMAE} & \(\textbf{NMaxE}\) $\downarrow$ \\\midrule$z=-0.7$    & xx     & xx  \\$z=-0.6$    & xx     & xx  \\$z=-0.5$    & xx     & xx  \\$z=-0.4$    & xx     & xx \\$z=-0.3$    & xx     & xx \\\bottomrule\end{tabular}\end{table}

\section{Notations}
\paragraph{General maths:}
\begin{itemize}
    \item $\Delta_\mathcal{S}$: Laplace Beltrami operator.
    \item $\Delta$: standard laplacian.
    \item $\nabla_\mathcal{S}$: surface gradient.
    \item $\nabla$: standard gradient.
    \item $\mathbb{S}^2$: unit sphere of dim 2.
    \item $\varepsilon$: band thickness.
    \item $\Delta x$: grid points spacing.
    \item $T$: temperature parameter for global aggregation.
    \item $dt$: infénitésimal time step element.
    \item $\langle\cdot,\cdot\rangle$: dot product.
    \item $\mathrm{dist}(y,\mathcal{S})$: distance from a point $y$ to a surface $\mathcal{S}$.
\end{itemize}

\paragraph{Method:}
\begin{itemize}
    \item $\mathcal{S}$: surface representation.
    \item $(\mathbf{n}(x), \mathbf{t}_1(x), \mathbf{t}_2(x))$: normal and principal curvature directions at $x\in \mathcal{S}$. $\mathbf{t}_1$ corresponds to the maximum curvature direction.
    \item $G$: cartesian grid with a grid points spacing $\Delta x$ not specified in the notation.
    \item $\mathcal{B}_\mathcal{S}$: band coordinates around $\mathcal{S}$ with tickness $\varepsilon$ not specified in notation.
    \item $U$: function define on the entire band coords.  $\tilde{U}$ is same but also constant along the normals. Can also add underscore $t$ to say function at time $t$ ($U_t$).
    \item $u^i\in \mathbb{R}^k$: function define on the local part $\mathcal{B}_i$ of the band coords. $\tilde{u}^i$ is same but also constant along the normals. Can also add underscore $t$ to say function at time $t$ ($u_t^i$).
    \item $k$: number of points in the local band. 
\end{itemize}

\paragraph{Patch:}
\begin{itemize}
    \item $p_i^c$: center of the patch i. $p_i^c \in \mathcal{S}$.
    \item $\mathcal{L}_i$: local frame (i.e, $\mathcal{L}_i = \left(p_i^c, \mathbf{n}(p_i^c), \mathbf{t}_1(p_i^c), \mathbf{t}_2(p_i^c)\right)$.
    \item $\mathcal{F}_i$: surface features of patch i. 
    \item $\hat{\mathcal{F}}_i$: surface features expressed in the local frame $\mathcal{L}_i$.
    \item $\mathcal{B}_i$: local band coordinates of patch i.
    \item $\hat{\mathcal{B}}_i$ local band coordinates expressed in the local frame $\mathcal{L}_i$.
    \item $\mathcal{P}$: patch or $\mathcal{P}_i$ if need to specify its index.
\end{itemize}

\paragraph{Network:}
\begin{itemize}
    \item $\mathcal{N}_\Theta: (q, \hat{\mathcal{B}}_i, \hat{\mathcal{F}}_i, u)\longmapsto\mathcal{N}_\Theta (q, \hat{\mathcal{B}}_i, \hat{\mathcal{F}}_i, u)$: global network.
    \item $\Phi_{\theta_1}$, $\Phi_{\theta_2}$, $\Phi_{\theta_3}$: MLPs.
    \item $\mathcal{N}_\Theta^{(\mathcal{P}, u)}:q\longmapsto\mathcal{N}_\Theta (q, \hat{\mathcal{B}}, \hat{\mathcal{F}}, u)$: compressed network version where the patch and function are fixed. $(\hat{\mathcal{B}}, \hat{\mathcal{F}})\in \mathcal{P}$.
    \item $\mathcal{N}_\Theta^{(\mathcal{P}, u)}(Q)=\left\{\mathcal{N}_\Theta^{(\mathcal{P}, u)}(q)\right\}_{q\in Q}$ where $Q\in R^{b\times3}$. Vectorised version of $\mathcal{N}_\Theta$.
\end{itemize}

\paragraph{Training pipeline:}
\begin{itemize}
    \item $\operatorname{cp}$: closest point operator.
    \item $\Pi_i$: image of $\mathcal{B}_i$ by $\operatorname{cp}$.
    \item $\mathcal{M}$: set of monomials.
    \item $g$: to designate a given monomial.
    \item $\mathcal{D}$: dataset.
    \item $\mathcal{E}_i$: contains the functions (input/GT) for training.
    \item $
    L_{\mathrm{NC}},L_{\mathrm{MSE}}$: losses.
\end{itemize}

\end{document}